\PassOptionsToPackage{table}{xcolor}
\documentclass{style}

\newif\ifrenderprompttemplates
\renderprompttemplatestrue
\newcommand{\prompttemplateref}[2]{\ifrenderprompttemplates\ref{#1}\else#2\fi}

\usepackage{amsmath,amsfonts,bm}

\def\eqref#1{equation~\ref{#1}}
\def\1{\bm{1}}

\DeclareMathAlphabet{\mathsfit}{\encodingdefault}{\sfdefault}{m}{sl}
\SetMathAlphabet{\mathsfit}{bold}{\encodingdefault}{\sfdefault}{bx}{n}

\usepackage{graphicx}
\usepackage{float}
\usepackage{eso-pic}
\usepackage{booktabs}
\usepackage{multirow}
\usepackage{siunitx}
\usepackage{capt-of}
\usepackage{colortbl}
\usepackage{placeins}
\usepackage{hyperref}
\usepackage{url}
\usepackage{tikz}
\usepackage{listings}
\usepackage{amssymb}
\usepackage{algorithm}
\usepackage{algpseudocode}
\newcommand{\genIcon}[1]{%
  \tikz[x=0.72em,y=0.72em,baseline=-0.08em,line width=0.42pt,
        line cap=round,line join=round]{%
    \path[use as bounding box] (0,0) rectangle (1,1);
    #1}}
\newcommand{\genInterleaved}{\genIcon{%
  \draw (0.05,0.52) rectangle (0.53,0.95);
  \draw (0.08,0.57)--(0.24,0.75)--(0.34,0.65)--(0.43,0.78)--(0.50,0.57);
  \draw (0.65,0.88)--(0.98,0.88) (0.65,0.66)--(0.98,0.66);
  \draw (0.05,0.32)--(0.98,0.32) (0.05,0.10)--(0.76,0.10);}}
\newcommand{\genGrid}{\genIcon{%
  \draw (0.05,0.05) rectangle (0.95,0.95);
  \draw (0.50,0.05)--(0.50,0.95) (0.05,0.50)--(0.95,0.50);}}

\newcommand{\genVideo}{\genIcon{%
  \draw (0.02,0.10) rectangle (0.98,0.90);
  \draw (0.20,0.10)--(0.20,0.90) (0.80,0.10)--(0.80,0.90);
  \foreach \y in {0.30,0.50,0.70}{
    \draw (0.02,\y)--(0.20,\y) (0.80,\y)--(0.98,\y);}
  \fill (0.39,0.29)--(0.39,0.71)--(0.67,0.50)--cycle;}}
\newcommand{\genPlan}{\genIcon{%
  \draw (0.06,0.08) rectangle (0.59,0.94);
  \foreach \y in {0.73,0.51,0.29}{
    \fill (0.17,\y) circle[radius=0.035];
    \draw (0.29,\y)--(0.48,\y);}
  \draw (0.69,0.76) .. controls (1.10,0.71) and (1.09,0.22) .. (0.70,0.19);
  \draw (0.81,0.07)--(0.69,0.19)--(0.81,0.31);}}

\newcommand{\worldcastmark}{\texorpdfstring{\raisebox{-0.0867\fontcharht\font`H}{\includegraphics[height=1.1443\fontcharht\font`H]{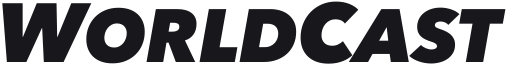}}\kern-.03em}{WorldCast}}
\ifcsname iflatexml\endcsname\else\expandafter\newif\csname iflatexml\endcsname\fi
\iflatexml\renewcommand{\worldcastmark}{WorldCast}\fi
\title{WorldCast: Exploring Intelligent Visual Generation in World State Prediction}
\newcommand{\benchmark}{\textsc{WorldCast}}
\definecolor{groupbestcolor}{RGB}{214,236,244}
\definecolor{globalbestcolor}{RGB}{252,218,185}
\definecolor{promptbackground}{RGB}{247,248,250}
\definecolor{promptborder}{RGB}{190,198,207}
\newcommand{\groupbest}[1]{\cellcolor{groupbestcolor}\textbf{#1}}
\newcommand{\globalbest}[1]{\cellcolor{globalbestcolor}\textbf{#1}}
\lstdefinestyle{prompt}{
  basicstyle=\ttfamily\fontsize{9}{11}\selectfont,
  backgroundcolor=\color{promptbackground},
  rulecolor=\color{promptborder},
  frame=single,
  framerule=0.45pt,
  breaklines=true,
  breakatwhitespace=true,
  columns=fullflexible,
  keepspaces=true,
  showstringspaces=false,
  literate={×}{{$\times$}}1 {—}{{\textemdash}}1,
  aboveskip=5pt,
  belowskip=7pt,
  xleftmargin=2pt,
  xrightmargin=2pt,
  framexleftmargin=4pt,
  framexrightmargin=4pt
}

\iflatexml
\author{Wenhao Yan\textsuperscript{1,*} \quad Haodong Li\textsuperscript{3} \quad Xiaotong Li\textsuperscript{2,\textdagger} \quad Yaxin Luo\textsuperscript{5} \quad Zhengrong Yue\textsuperscript{6} \quad Jialv Zou\textsuperscript{7} \quad Xu Huang\textsuperscript{4} \quad Haojin Yang\textsuperscript{4} \quad Nan Jia\textsuperscript{7} \quad Yan Feng\textsuperscript{2} \quad Peng Pei\textsuperscript{2} \quad Xunliang Cai\textsuperscript{2}
\\ \textsuperscript{1}THU \quad \textsuperscript{2}Meituan LongCat \quad \textsuperscript{3}CUHK \quad \textsuperscript{4}PKU \quad \textsuperscript{5}MBZUAI \quad \textsuperscript{6}SJTU \quad \textsuperscript{7}HUST
\\ {\small\textsuperscript{*}Work done during internship at Meituan.\quad\textsuperscript{\textdagger}Project Lead.}
\\ \textbf{Project Page:} \url{https://teal024.github.io/WorldCast}}
\else
\author[1,*]{Wenhao Yan}
\author[3]{Haodong Li}
\author[2,\textdagger]{Xiaotong Li}
\author[5]{Yaxin Luo}
\author[6]{Zhengrong Yue}
\author[7]{Jialv Zou}
\author[4]{Xu Huang}
\author[4]{Haojin Yang}
\author[7]{Nan Jia}
\author[2]{Yan Feng}
\author[2]{Peng Pei}
\author[2]{Xunliang Cai}
\affiliation[1]{THU}
\affiliation[2]{Meituan LongCat}
\affiliation[3]{CUHK}
\affiliation[4]{PKU}
\affiliation[5]{MBZUAI}
\affiliation[6]{SJTU}
\affiliation[7]{HUST}
\contribution{\fontsize{7}{8.5}\selectfont\textsuperscript{*}Work done during internship at Meituan.\quad\textsuperscript{\textdagger}Project Lead.}

\renewcommand{\projectpageline}{\textbf{Project Page:} \url{https://teal024.github.io/WorldCast}}
\fi

\abstract{
Humans interpret multimodal observations of the world and forecast how it will evolve. Recent multimodal generative systems have demonstrated similar capabilities: they can integrate multimodal context and synthesize visual outputs that depict changes in the world. To characterize this emerging trajectory, we organize model capabilities into a four-level hierarchy and discover that higher-level tasks remain relatively underexplored despite model progress.
To bridge this gap, we explore model performance on multi-step world state prediction, a high-level task that requires models to forecast successive future world states from multimodal observations and diverse references. Such a task requires models to possess entangled understanding and generation capabilities, as well as extensive world knowledge to translate human action sequences, physical transformations, and abstract visual reasoning into coherent visual outputs. We introduce \benchmark{}, a benchmark comprising 320 manually designed cases across eight categories, and evaluate 12 representative models.
The conducted experiments yield three key insights: (a) advances in frontier models are rapidly pushing the boundaries of intelligent world modeling, with the GPT-Image-2.5 System scoring 72.8 and demonstrating early-stage visual state prediction capabilities on real-world tasks; (b) \benchmark{} reveals substantial gaps between models: even Nano Banana Pro reaches only 42.8 and the best open-source model scores 20.1, with deficits in handling complex context, entangled references, and reasoning-heavy tasks; (c) strong generative models effectively draw on the capabilities of their underlying language models. These findings highlight the need for multimodal generative models that are capable of dynamic planning and reasoning while also being deeply unified to faithfully simulate real-world states.}

\begin{document}

% First-page wordmark in the upper-left margin, as in the LongCat report.
\AddToShipoutPictureFG*{%
  \AtPageUpperLeft{%
    \raisebox{-1.97cm}[0pt][0pt]{%
      \hspace{2.5cm}\includegraphics[height=20.5pt]{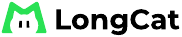}}}}
\maketitle

\vspace{4pt}
\begingroup
\setlength{\intextsep}{3pt}
\begin{figure}[H]
    \centering
  \includegraphics[width=\linewidth]{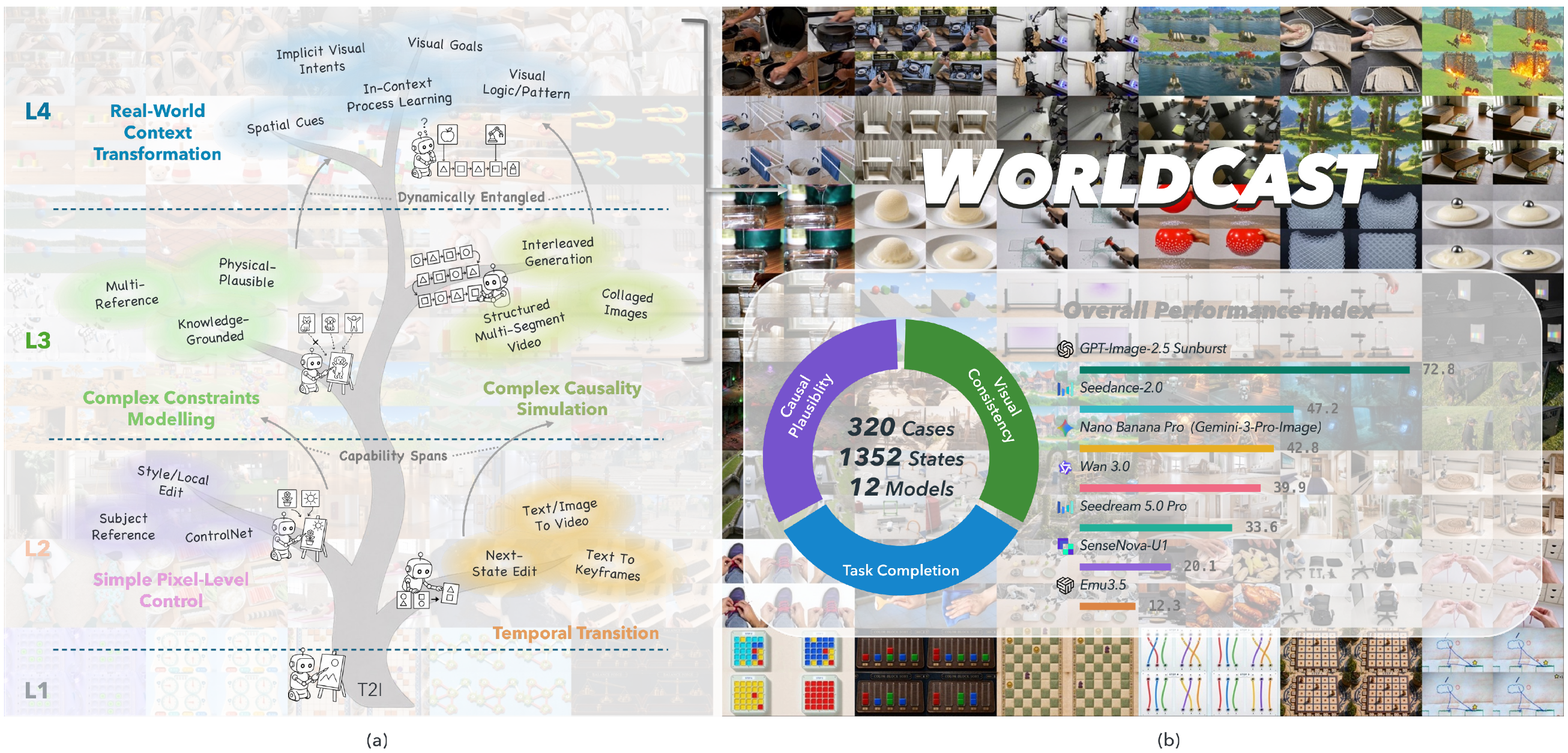}\par
  \setlength{\abovecaptionskip}{6pt}
  \caption{Our generative capability tree and the \benchmark{} benchmark. (a) The proposed visual generative capability tree organizes intelligent visual generation into a four-layer hierarchy (L1-L4); (b) \benchmark{} targets L3-L4 level capabilities of dynamically transforming real-world context into successive future world states through unified multimodal understanding and generation.}
  \label{fig:capability-tree}
\end{figure}
\endgroup

\begin{figure}[!t]
\centering
\includegraphics[width=\linewidth]{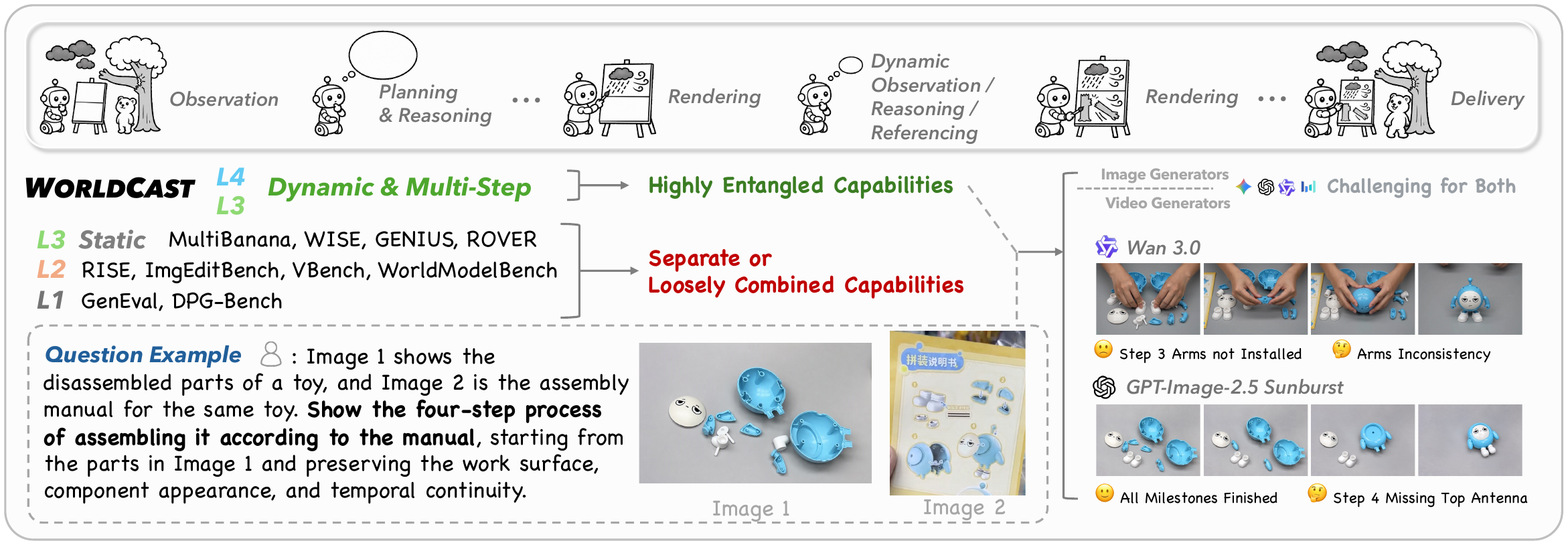}
\setlength{\abovecaptionskip}{6pt}
\caption{Evaluation scope and question example of \benchmark{}. The benchmark requires models to combine understanding and generation to dynamically produce evolving visual states. In this assembling task, GPT-Image-2.5 achieves key milestones while Wan 3.0 struggles in figuring out the correct order.}
\label{fig:world-prediction-example}
\end{figure}
\section{Introduction}

Multimodal generative systems have advanced rapidly on both the understanding and generation sides: they can understand increasingly rich inputs and produce
increasingly expressive visual outputs. Recent image and video generation
systems can combine language with visual context ~\citep{google2025nanobananapro, openai2025gptimage}, understand and utilize referenced content,
and render multi-stage changes rather than isolated static scenes
\citep{kuaishou2026kling3,seedance2026seedance2}. As with human intelligence,
once these capabilities reach a sufficient level, they can in turn benefit real-world tasks \citep{zhang2026imagewam}. This
trajectory resembles how humans learn and reason: we observe a multimodal world,
draw upon relevant multimodal cues when confronting a problem, and
envision complex ways in which the world may evolve.

To trace the progression of model capabilities, we organize them into four levels, illustrated in
Figure~\ref{fig:capability-tree}. Basic text-to-image (T2I) ability focuses on a static generation process; editing, controllable generation, and video generation extend these capabilities to L2. As frontier models advance in both understanding and generation \citep{google2025nanobananapro, openai2025sora2, seedream2025seedream45}, they support interleaved or segmented generation and heterogeneous references, and the generation process itself becomes more intelligent by drawing on the strengths of powerful language models\citep{openai2026gpt5.5, anthropic2025claudeopus4}.

Despite rapid progress, higher-level model capabilities that require tightly integrating understanding and generation remain largely unexplored. Such capabilities, however, are highly relevant to the modeling of real-world tasks. Prior works mainly explore whether models can simulate our world from static appearance anchors \citep{videoasprompt, omnivbench}, yet real-world events are mostly dynamic and predicting their progression requires references to intervene at different temporal steps. Those references may specify a state to observe, a goal, rules, patterns, object properties or relevant prior experience, rather than a starting point or appearance signals.
We refer to this direction as \emph{\textbf{world state prediction}}: \emph{utilizing heterogeneous context to guide an evolving real-world process}. Such a task requires a model not only to output successive visual states but also to reason about what each reference means and when and how it should take effect.

To explore model performance in this direction, we propose \benchmark{}, a comprehensive benchmark designed to answer the question: can multimodal generative systems externalize L3-L4 level intelligence in world state prediction? \benchmark{} targets high-level simulation of real-world tasks, hierarchically distinguished by how deeply and dynamically a model must integrate reference understanding with control and knowledge. We collect L3-level cases focusing on complex dynamic transitions and physical plausibility, which require models to engage in implicit or explicit planning and reasoning. L4-level cases are designed for models to integrate references that
may encode goals, physical states, materials, experience or patterns and translate them into a unified understanding that guides complex state transitions. The latter tasks provide richer contextual information yet place greater demands on tightly integrated multimodal intelligence.

To accommodate both image and video models, we evaluate all model outputs as ordered sequences of keyframes: regardless of whether generation is in a grid, interleaved, or in a continuous sequence, its success must ultimately be reflected in the resulting visual states. Based on the protocol, we further introduce manually annotated case-specific hints for reliable sequence-level LMM-as-a-judge evaluation. The final benchmark evaluates 12 models in 320 test cases across 8 categories.

Our experiments show that frontier unified models have reached an early stage
of intelligent world simulation: they can successfully handle some L3 and even L4 tasks and show emerging generalization across tasks, yet some domains remain challenging. We further find that interleaved generation improves the performance of frontier image models, enabling them to capture critical temporal transitions without relying on video architectures. We observe that the substantial model performance gaps are closely tied to their language-side capabilities. Building on this finding, we combine visual generation with agentic capabilities, including inspection-revision loops and coding tools, yielding a clearer understanding of their benefits and suggesting potential research paths toward more capable unified generation systems.

In summary, our contributions are threefold:

\noindent \textbf{(1)} To systematically investigate the intelligence of visual generative systems, we organize visual generative capabilities into a four-level hierarchy toward real-world simulation and contribute an early effort to formalize the  \emph{\textbf{world state prediction}} task. The task focuses on L3--L4 capabilities of translating heterogeneous contexts into successive future world states, which remains insufficiently explored by existing benchmarks.

% \noindent \textbf{(2)} \benchmark{} is elaborately tailored for evaluating such task, comprising 320 manually curated cases across 8 task categories and 1,352 requested visual states that comprehensively assess the intertwined capabilities of visual understanding and generation. A total of 12 advanced models spanning both image and video generation are evaluated under a unified state-sequence protocol, with performance measured across three dimensions using case-specific annotations.

\noindent \textbf{(2)} \benchmark{} offers a benchmark tailored to evaluating these capabilities, comprising 320 manually curated cases and 1,352 requested visual states across 8 task categories that comprehensively assess the intertwined capabilities of visual understanding and generation. A total of 12 advanced models spanning both image and video generation are evaluated under a unified state-sequence protocol across three dimensions.

\noindent \textbf{(3)} A systematic diagnosis with \benchmark{} exposes widening gaps among current generative systems at higher levels and identifies key factors behind model performance on this task. Further analyses of \emph{interleaved generation}, \emph{language enhancement}, and \emph{agentic behavior} suggest promising directions while highlighting the substantial challenges of simulating real-world states and achieving advanced generative intelligence.

\section{\worldcastmark}
\benchmark{} is grounded in the cognitively motivated capability hierarchy introduced above. We distinguish task categories by where the information governing a future process is encoded and what reasoning is required to translate it into visual state transitions. The L3--L4 distinction reflects how tightly reference understanding and transition generation must be integrated with additional context, which may make a task easier for humans while demanding more integrated multimodal reasoning from models.

To simulate real-world tasks, \benchmark{} requires models to output causally coherent world states from heterogeneous references. Given text instructions $p$ and one or more context images $\mathcal{X}=(x_m)_{m=1}^{M}$, the generative system $G$ predicts an ordered visual sequence:
\begin{equation}
  \widehat{\mathcal{Y}} = G(p,\mathcal{X})
  = (\hat{y}_1,\ldots,\hat{y}_K),
  \label{eq:worldcast-formulation}
\end{equation}
where $\widehat{\mathcal{Y}}$ denotes the predicted sequence of $K$ visual world states.
As illustrated in Figure~\ref{fig:benchmark-overview}, \benchmark{} comprises eight task categories across two levels under this formulation:

\begin{figure}[!t]
  \centering
  \includegraphics[width=\linewidth]{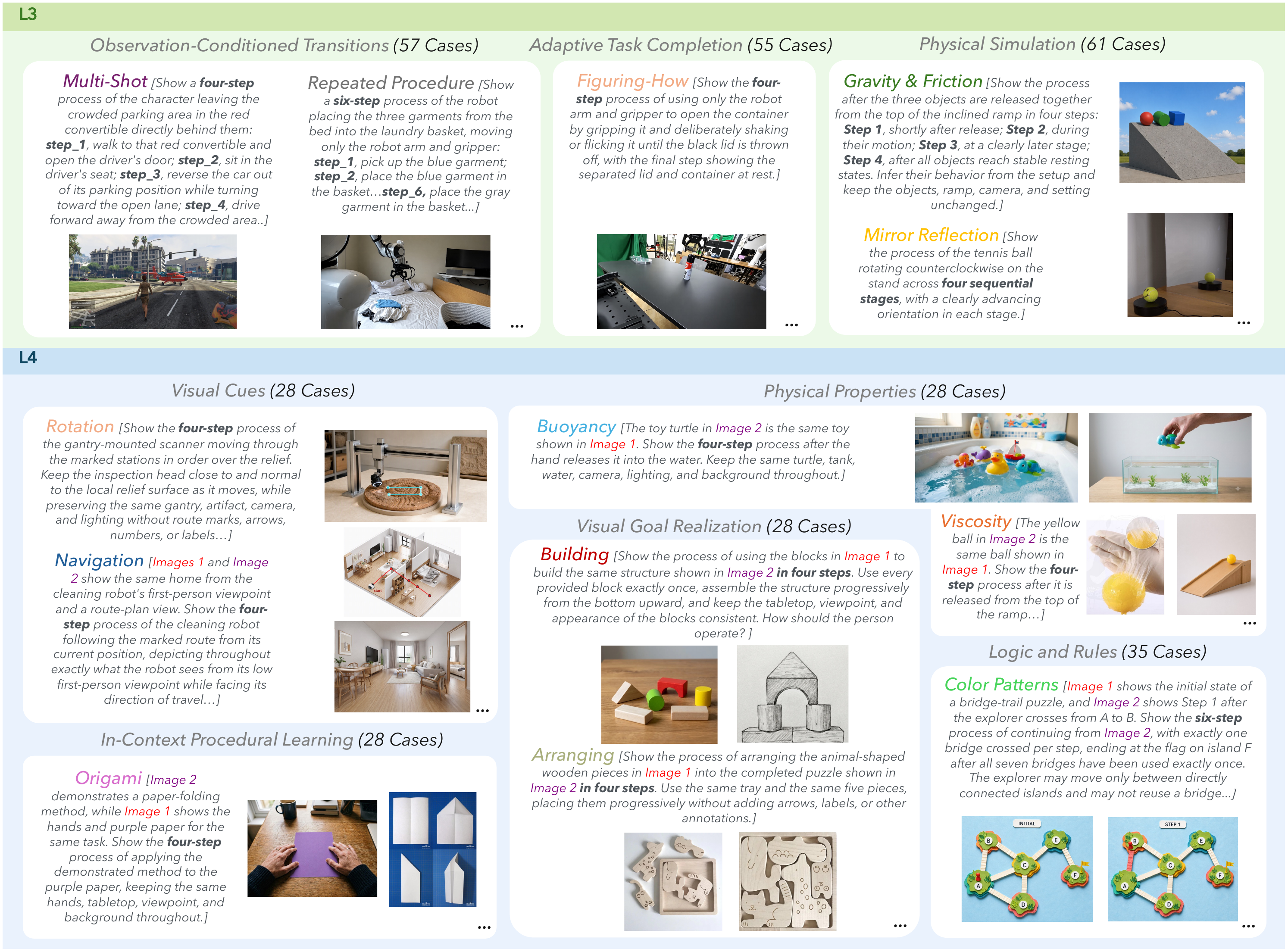}\par
  \setlength{\abovecaptionskip}{6pt}
  \caption{Representative examples of the eight \benchmark{}
  categories. The benchmark comprises 320 manually designed cases in total, with 1,352 requested visual states.}
  \label{fig:benchmark-overview}
  \end{figure}

\begingroup
\setlength{\parskip}{2pt}
\noindent\textbullet\ \textbf{Observation-Conditioned Transitions (L3).}
Execution and rendering require observing the initial world rather than simply moving around. Step-level instructions force the model to ground each action in the observed world.
\par\noindent\textbullet\ \textbf{Physical Simulation (L3).}
Models should depict how an event unfolds from the initial state according to physical laws or real-world dynamics, connecting observations with physical knowledge.
\par\noindent\textbullet\ \textbf{Adaptive Task Completion (L3).}
Models must decompose the goal without step-level instructions, infer feasible intermediate actions, and ground their plans in the changing visual state.
\par\noindent\textbullet\ \textbf{Visual Cues (L4).}
Text alone is sometimes insufficient to specify location or transitions. Direct or indirect visual cues combined with text instructions indicate where or how the world should change.
\par\noindent\textbullet\ \textbf{Physical Properties (L4).}
Observations may include material properties and object states without fully verbalizing them. The model must infer
the operative property, bind it to the correct entities, and carry its consequences through a new multi-state simulation.
\par\noindent\textbullet\ \textbf{Visual Goal Realization (L4).}
The desired outcome is specified visually, either as a final state or hints of the final state. The model must identify the goal semantics, plan how to achieve the final state, and synthesize the intervening steps.
\par\noindent\textbullet\ \textbf{Logic and Rules (L4).}
Several historical states and textual cues jointly define a
transition rule. The model must combine textual reasoning with
visual rules, induce the operation for successive states and render it consistently.

\par\noindent\textbullet\ \textbf{In-Context Procedural Learning (L4).}
Text and images describe prior experience in the form of an
illustrated procedure. The model must identify the function of each step,
abstract the procedure from its original scene, and transfer it to a new world
rather than imitate the examples superficially.
\par\endgroup

\subsection{Evaluation Metrics}

\subsubsection{Evaluation Dimensions}

We evaluate each generated sequence along three complementary dimensions: Task
Completion (TC), Causal Plausibility (CP), and Visual Consistency (VC).

\noindent\textbf{Task Completion (TC): } TC measures whether the sequence visibly realizes the user's intended task. It is computed as a weighted average of the sequence-level sub-metric \textbf{Progress Alignment (PA)} and the outcome-level sub-metric \textbf{Outcome Completion (OC)}. Outcome weight ranges from 0 to 0.5 for different cases; at zero, only PA is evaluated. Calculation details are in Appendix~\ref{app:evaluation-rubrics}.

\noindent\textbf{Causal Plausibility (CP): }
CP evaluates whether adjacent transitions and longer-range dependencies follow
the applicable physical laws, logical rules, or task-specific world dynamics.
It penalizes unsupported changes such as impossible transformations or reversed
cause and effect.

\noindent\textbf{Visual Consistency (VC): }
VC evaluates whether visual properties that should persist---including object
identity, count, geometry, scale, material, camera, and background---remain
stable across the sequence. Changes required by the task are exempt; unrelated
appearance, disappearance, duplication, deformation, or drift is penalized.

We aggregate TC, CP, and VC derived from discrete ratings on a 0--3 scale,
then convert the result to a 100-point score:
\begin{equation}
  \mathrm{Final} = \operatorname{Rescale}(\alpha_1\,\mathrm{TC} + \alpha_2\,\mathrm{CP}
  + \alpha_3\,\mathrm{VC}).
\end{equation}

\begin{figure}[!t]
  \centering
  \includegraphics[width=\linewidth]{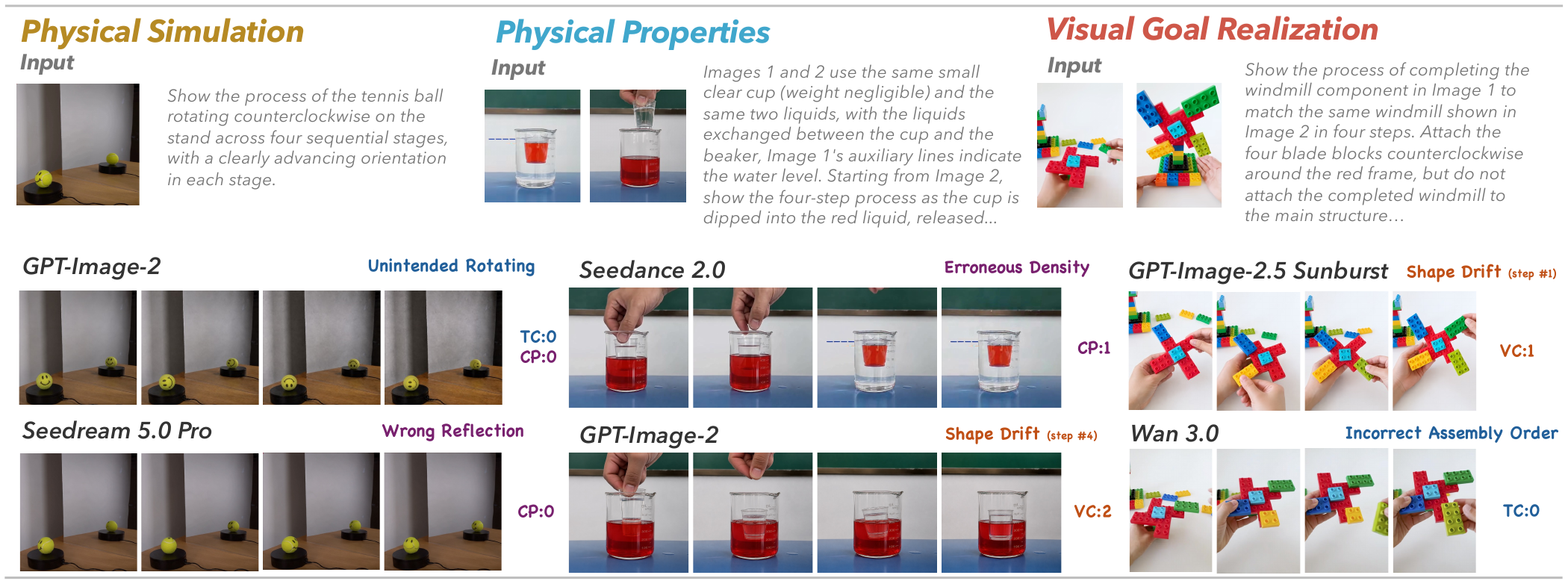}
  \caption{Qualitative examples on \benchmark{}. Models exhibit distinct failure patterns across the three evaluation dimensions; all scores are reported on a 0--3 scale.}
  \label{fig:qualitative-comparison}
\end{figure}

\section{Evaluation Protocol}
\label{sec:evaluation-protocol}

\par\noindent\textbf{Agentic Video Adapter.}\quad
To evaluate image and video models under the same sequence-level protocol, we
introduce an agentic video adapter that converts each generated video into an
ordered set of task-aligned keyframes. The adapter follows an inclusion-first
principle, selecting frames that cover as many critical states as
possible while preserving their temporal constraints. In practice, we adopt GPT-6 Astra~\citep{openai2026gpt6} through Codex~\citep{openai2025codex} as the agentic video adapter.  The adapter is used only to determine whether the video contains the required key states; it does not assess the plausibility of the full video sequence or its visual quality.

\par\noindent\textbf{LMM-as-a-Judge.}\quad
We use GPT-5.5 as the automatic evaluator. Given the task instruction,
multimodal references, judge-only hints, and the ordered output sequence, the
judge scores Progress Alignment (PA) and Outcome Completion (OC) to compute
Task Completion (TC), together with Causal Plausibility (CP) and Visual
Consistency (VC). The same evaluation
protocol is applied to image-model outputs and video keyframes selected by the
agentic video adapter. Full scoring criteria and aggregation rules are provided
in Appendix~\ref{app:evaluation-rubrics}.

\noindent\textbf{Human-Annotated Hints for Judges.}\quad
Because we evaluate entire output sequences and some tasks are partially open-ended, judging can involve substantial reasoning. To improve evaluation consistency and accuracy, we introduce task-specific judge-only hints. Human annotators provide manual hints that identify task-critical objects, milestones, and constraints, and highlight deterministic steps the judge should not overlook. Appendix~\ref{app:human-hints} provides further details.

\section{Experiments}

\begin{table}[t]
  \centering
  \small
  \setlength{\tabcolsep}{2.8pt}
  \setlength{\abovecaptionskip}{6pt}
  \setlength{\belowcaptionskip}{5pt}
  \caption{Overview of \benchmark{} performance across twelve image and video
  generation models and eight benchmark categories. Blue cells mark the best
  result within each model group, while orange cells mark the best result across
  all models. Icons indicate the generation format:
  \protect\genInterleaved\ Interleaved Generation,
  \protect\genGrid\ Grid Generation,
  \protect\genVideo\ Video Generation,
  \protect\genPlan\ Plan \& Iterate.}
  \label{tab:main-results}
  \renewcommand{\arraystretch}{1.5}
  \resizebox{0.97\linewidth}{!}{%
  \begin{tabular}{>{\fontsize{10}{12}\selectfont\raggedright\arraybackslash}m{3.0cm}|
                  *{3}{>{\fontsize{11}{13}\selectfont\centering\arraybackslash}m{1.7cm}}|
                  >{\fontsize{11}{13}\selectfont\centering\arraybackslash}m{1.7cm}|
                  *{5}{>{\fontsize{11}{13}\selectfont\centering\arraybackslash}m{1.7cm}}|
                  >{\fontsize{11}{13}\selectfont\centering\arraybackslash}m{1.7cm}|
                  >{\fontsize{11}{13}\selectfont\centering\arraybackslash}m{1.7cm}}
    \specialrule{0.8pt}{0pt}{0pt}
    \multirow{2}{*}[-8pt]{\fontsize{11}{13}\selectfont\textbf{Method}}
    & \multicolumn{4}{c|}{\fontsize{11}{13}\selectfont\textbf{L3 Tasks}}
    & \multicolumn{6}{c|}{\fontsize{11}{13}\selectfont\textbf{L4 Tasks}}
    & \multirow{2}{*}[-8pt]{\fontsize{11}{13}\selectfont\textbf{Overall}} \\
    \cmidrule(lr){2-5}\cmidrule(lr){6-11}
    & \fontsize{9.5}{11}\selectfont\itshape\shortstack[c]{Observation-\\Conditioned\\Transitions}
    & \fontsize{9.5}{11}\selectfont\itshape\shortstack[c]{Physical\\Simulation}
    & \fontsize{9.5}{11}\selectfont\itshape\shortstack[c]{Adaptive\\Task\\Completion}
    & \fontsize{9}{10.5}\selectfont\textbf{\shortstack[c]{L3\\Overall}}
    & \fontsize{9.5}{11}\selectfont\itshape\shortstack[c]{Physical\\Properties}
    & \fontsize{9.5}{11}\selectfont\itshape\shortstack[c]{Visual\\Cues}
    & \fontsize{9.5}{11}\selectfont\itshape\shortstack[c]{In-Context\\Procedural\\Learning}
    & \fontsize{9.5}{11}\selectfont\itshape\shortstack[c]{Logic and\\Rules}
    & \fontsize{9.5}{11}\selectfont\itshape\shortstack[c]{Visual Goal\\Realization}
    & \fontsize{9}{10.5}\selectfont\textbf{\shortstack[c]{L4\\Overall}}
    & \\
    \specialrule{0.5pt}{0pt}{0pt}
    \rowcolor{black!7}
    \multicolumn{12}{c}{\fontsize{10}{12}\selectfont\textbf{\textit{Open-source Image Models}}} \\
    SenseNova-U1~\genInterleaved & \groupbest{24.1} & \groupbest{24.6} & \groupbest{24.7} & \groupbest{24.4} & \groupbest{19.9} & \groupbest{19.1} & \groupbest{20.1} & \groupbest{5.9} & \groupbest{12.2} & \groupbest{15.0} & \groupbest{20.1} \\
    Emu3.5~\genInterleaved & 17.4 & 13.8 & 18.5 & 16.5 & 6.6 & 10.3 & 8.1 & 4.8 & 7.6 & 7.4 & 12.3 \\
    BAGEL~\genPlan & 4.9 & 7.0 & 8.3 & 6.7 & 8.5 & 12.5 & 7.0 & 2.9 & 4.0 & 6.8 & 6.8 \\
    Qwen-Image~\genGrid & 7.8 & 6.8 & 6.2 & 6.9 & 4.0 & 6.9 & 6.1 & 0.6 & 3.4 & 4.0 & 5.6 \\
    \rowcolor{black!7}
    \multicolumn{12}{c}{\fontsize{10}{12}\selectfont\textbf{\textit{Proprietary Image Systems}}} \\
    GPT-Image-2.5~\genInterleaved & \globalbest{85.6} & 69.6 & \globalbest{66.1} & \globalbest{73.8} & \globalbest{73.6} & \globalbest{76.7} & \globalbest{71.6} & \globalbest{74.3} & \globalbest{61.3} & \globalbest{71.6} & \globalbest{72.8} \\
    GPT-Image-2~\genInterleaved & 74.2 & \globalbest{70.6} & 59.6 & 68.3 & 66.0 & 62.6 & 70.6 & 66.6 & 51.8 & 63.7 & 66.2 \\
    Nano Banana Pro~\genInterleaved & 36.2 & 56.1 & 43.5 & 45.6 & 49.9 & 33.7 & 43.3 & 41.4 & 28.5 & 39.5 & 42.8 \\
    Seedream 5.0 Pro~\genGrid & 30.3 & 45.9 & 33.4 & 36.8 & 46.7 & 27.7 & 38.9 & 11.9 & 29.1 & 29.9 & 33.6 \\
    Seedream 4.5~\genGrid & 19.4 & 14.9 & 18.7 & 17.6 & 19.7 & 16.0 & 21.4 & 3.4 & 12.8 & 14.1 & 16.0 \\
    \rowcolor{black!7}
    \multicolumn{12}{c}{\fontsize{10}{12}\selectfont\textbf{\textit{Video Systems}}} \\
    Seedance 2.0~\genVideo & \groupbest{53.3} & \groupbest{53.4} & \groupbest{48.7} & \groupbest{51.9} & 40.8 & \groupbest{56.6} & \groupbest{55.4} & \groupbest{17.7} & \groupbest{43.3} & \groupbest{41.6} & \groupbest{47.2} \\
    Gemini Omni~\genVideo & 45.6 & 50.2 & 41.0 & 45.8 & \groupbest{47.2} & 42.7 & 49.9 & 16.2 & 33.0 & 36.8 & 41.6 \\
    Wan 3.0~\genVideo & 51.9 & 47.3 & 38.5 & 46.0 & 42.7 & 42.7 & 35.5 & 17.2 & 29.6 & 32.8 & 39.9 \\
    \specialrule{0.8pt}{0pt}{0pt}
  \end{tabular}%
  }
\end{table}

\subsection{Settings}
We evaluate twelve representative image and video generation models across all eight task categories in \benchmark{}. The evaluated models are organized into three groups:

\noindent\textbullet\ \textbf{Open-source Image Models.} The suite comprises unified models
Emu3.5~\citep{cui2025emu35}, BAGEL~\citep{deng2025bagel}, SenseNova-U1~\citep{diao2026sensenovau1} and an image diffusion model supporting multi-reference generation, Qwen-Image~\citep{wu2025qwenimage}.

\noindent\textbullet\ \textbf{Proprietary Image Systems.} Proprietary image models include GPT-Image-2~\citep{openai2026gptimage2},
GPT-Image-2.5 Sunburst~\mbox{\citep{openai2026gptimage25sunburst}},
Nano Banana Pro~\citep{google2025nanobananapro},
Seedream 4.5~\citep{seedream2025seedream45}, and
Seedream 5.0 Pro~\citep{seedream2026seedream50pro}. Unless otherwise specified, GPT-Image-2 refers to interleaved generation through the Responses API paired with GPT-5.5~\citep{openai2026gpt5.5}, while GPT-Image-2.5 refers to the Sunburst version paired with GPT-6 Astra~\citep{openai2026gpt6}. Both language models were the respective flagship models at the time of the image models' releases.

\noindent\textbullet\ \textbf{Video Systems.}
We additionally evaluate three video
generation systems: Seedance 2.0~\citep{seedance2026seedance2},
Wan 3.0~\citep{alibaba2026wan3}, and Gemini Omni~\citep{google2026geminiomni}. 

Because \benchmark{} requires an ordered sequence of visual states, we adapt the inference format to each model's native output interface. We provide generation templates for each output mode in Appendix~\prompttemplateref{app:prompt-templates}{I}: image models or systems with interleaved generation capabilities produce mixed sequences of text and visuals, and the final performance in evaluated only on the visual states. Models or systems without interleaved capability, including Qwen-Image and Seedream, produce a visual grid, and evaluation is performed on the individual states cropped from that grid. For BAGEL, we adopt the Plan \& Iterate scheme following UEval~\citep{li2026ueval}. Video models are evaluated based on key world states extracted by the \textit{Agentic Video Adapter}.

As modern image and video generation systems increasingly integrate multiple components, such as LLM-based prompt enhancers~\citep{seedream2025seedream40,wanPE}, \benchmark{} evaluates overall multimodal generative intelligence through visual outputs, regardless of whether they are produced by a composite system, a single image or video model, or a unified multimodal generation model.
\subsection{Main Results}

\begin{figure}[H]
  \centering
  \setlength{\abovecaptionskip}{6pt}
  \includegraphics[width=\linewidth]{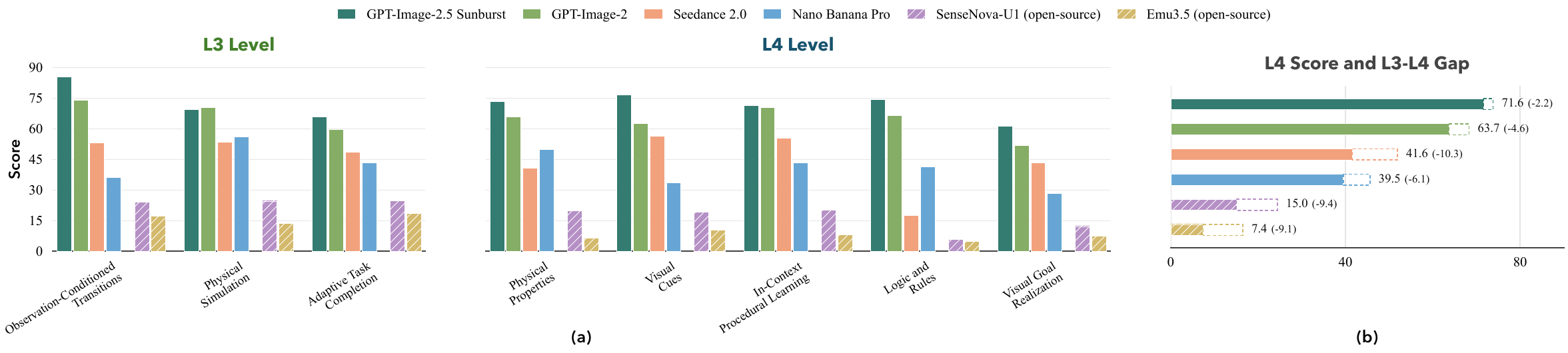}
  \caption{Performance comparison on \benchmark{}. (a) Category-wise scores across L3 and L4. (b) L4 scores, with dashed extensions and labels showing the absolute gap from L3.}
  \label{fig:level-comparison}
\end{figure}

\par\noindent\textbf{Frontier unified systems exhibit early-stage L4-level visual state prediction capabilities.}\quad
As shown in Table~\ref{tab:main-results}, the strongest proprietary image system GPT-Image-2.5 performs well on L3 tasks and scores above 60 on all five L4 tasks, indicating that frontier models already reveal an emerging ability to simulate how a world should evolve based on contextual information, and this ability is evolving rapidly. GPT-Image-2.5 can faithfully simulate the process of solving physical-world tasks in many household activities and scientific experiments, while outperforming other baselines in predicting evolution of digital games or solutions of puzzles. Nevertheless, even the strongest model does not consistently succeed across categories, achieving only 61.3 in the task of Visual Goal Realization, indicating that context-grounded world prediction remains a capability that has yet to be fully mastered by generative systems.

\par\noindent\textbf{\benchmark{} reveals substantial performance gaps between models that widen at higher levels.}\quad
The best open-source system trails substantially behind the strongest proprietary model in every category, and the gap generally widens on L4 tasks. For example, SenseNova-U1, the best-performing open-source model, drops from 24.4 on L3 to 15.0 on L4, whereas GPT-Image-2.5 declines by only 2.2 points, maintaining a score of 71.6 on L4.
The strongest model even performs better on some complex tasks, whereas weaker models struggle. For example, GPT-Image-2.5 scores 76.7 on Visual Cues and 74.3 on Logic and Rules, whereas Nano Banana Pro scores just 33.7 and 41.4, respectively, both below its overall score. This pattern suggests that, beyond a certain level of unified capability, models may benefit from the additional information provided by more complex inputs rather than be confused by it. The generation-only Qwen-Image performs worst, suggesting a need for understanding and reasoning in this benchmark.
Taken together, these results suggest that strong performance on \benchmark{} requires broad knowledge and generalizable generative intelligence.

\par\noindent\textbf{Performance varies across task categories but generally degrades on reasoning-intensive tasks.}\quad
% 语言reasoning 视觉reasoning
Models exhibit different weaknesses across tasks. Video models struggle
particularly with L4 Logic \& Rules, where digital simulation requires fine-grained logical reasoning and precise local changes.
Within L3, both GPT-Image models, Wan 3.0, and Gemini Omni
score lowest on Adaptive Task Completion that involves dynamic planning and execution. Within L4, Visual Goal Realization, where models have to extract actionable guidance from abstract or concrete visual goals, is the weakest category for both GPT-Image 2.5 and Nano Banana Pro, with the former achieving 61.3 and the latter only 28.5.
These patterns highlight the difficulty of aligning generation with temporal reasoning and abstract information.

\par\noindent\textbf{Visual continuity does not necessarily translate into accurate key states.}\quad
When tasks are evaluated through a sequence of semantically meaningful states, frontier image models capable of interleaved generation are surprisingly competitive with video models. The GPT-Image variants outperform Seedance 2.0 although the later scores highest among other remaining models. Since interleaved generation also supports training on video sequences, this raises the possibility that these image models benefit from both high-quality video training data to receive temporal supervision and linguistic reasoning that enhances task modeling.

\subsection{Discussion and Analysis}
\subsubsection{Analysis of the language model enhancer}
\label{sec:language-enhancer}

\begin{figure}[H]
  \centering
  \setlength{\abovecaptionskip}{6pt}
  \begin{minipage}[t]{0.5824\linewidth}
    \centering
    \includegraphics[width=0.92\linewidth]{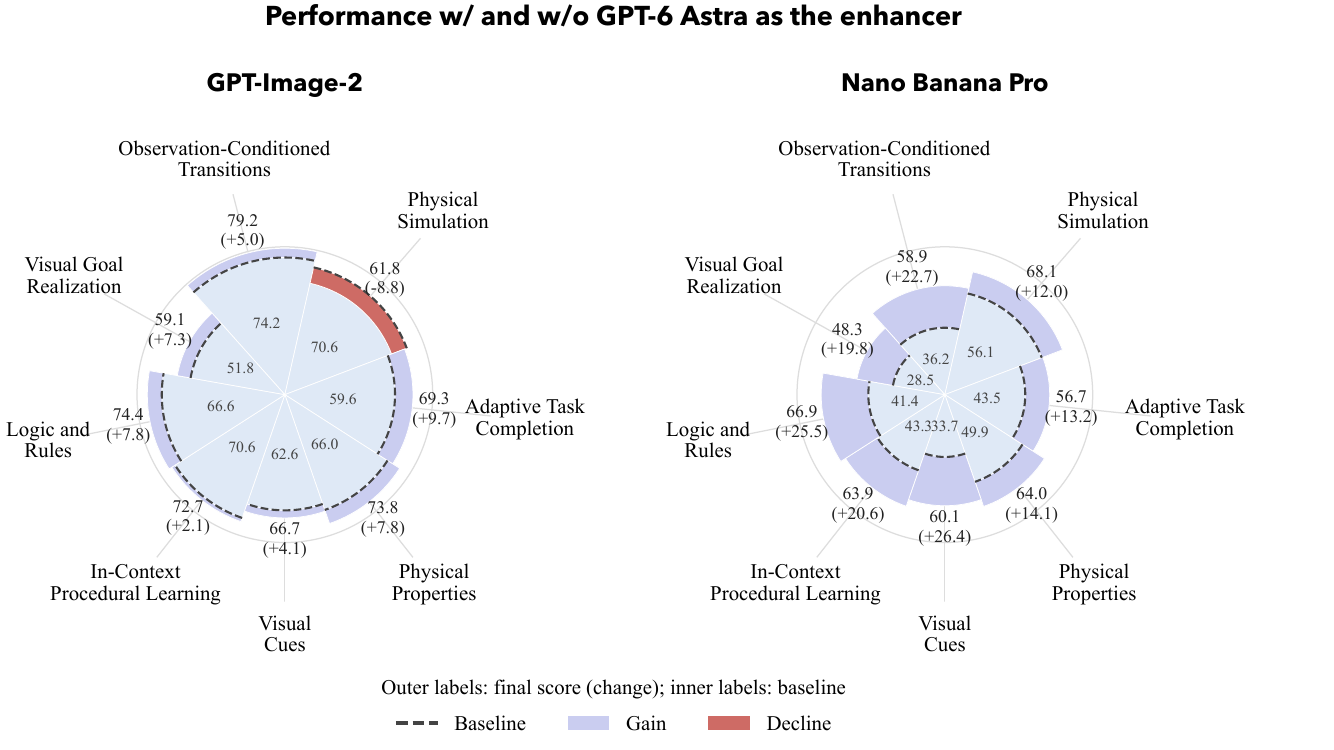}
    \caption{Category-wise scores and gains with GPT-6 Astra as the enhancer for two generators.}
    \label{fig:language-enhancement}
  \end{minipage}\hfill
  \begin{minipage}[t]{0.3976\linewidth}
    \centering
    \includegraphics[width=0.92\linewidth]{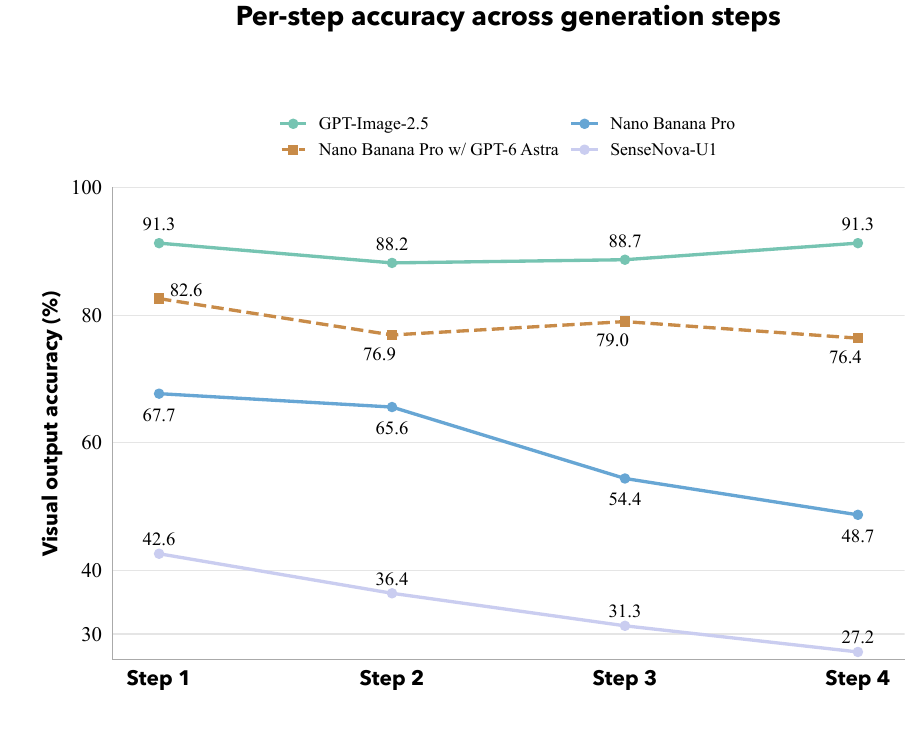}
    \caption{Visual output accuracy across four generation steps.}
    \label{fig:context-steps}
  \end{minipage}
\end{figure}

The main results show that the GPT-Image family substantially outperforms other model families on this benchmark. One plausible explanation is the contribution of the paired language model, whose extensive world knowledge and understanding capabilities enable it to produce enhanced, information-dense prompts. For analysis, we integrate Nano Banana Pro with GPT-6 and find that this combination substantially improves performance across L3-L4 task categories but still lags behind the GPT-Image family with gains varying across tasks, as shown in Figure~\ref{fig:language-enhancement}. When we replace GPT-5.5 with GPT-6 as the enhancer for the previous-generation GPT-Image-2, the overall score increases marginally with a decline in Physical Simulation. These results highlight the importance of the intelligent enhancer with broad knowledge, while also suggesting that effective guidance in world state prediction requires a strong visual generator and a closely matched enhancer.

\subsubsection{Analysis of interleaved generation}
\label{sec:generation-schema-analysis}

\begin{table}[H]
  \centering
  \small
  \setlength{\tabcolsep}{2.1pt}
  \setlength{\abovecaptionskip}{6pt}
  \setlength{\belowcaptionskip}{5pt}
  \caption{Interleaved versus grid generation. Switching to grid generation reduces overall performance for both models, with a much larger drop for GPT-Image-2.5.}
  \label{tab:generation-schema}
  \renewcommand{\arraystretch}{1.00}
  \resizebox{\linewidth}{!}{%
  \begin{tabular}{l|c|c|*{8}{c}|c}
    \toprule
    \textbf{Model}
    & \begin{tabular}[c]{@{}c@{}}\textbf{Generation}\\\textbf{Schema}\end{tabular}
    & \begin{tabular}[c]{@{}c@{}}\textbf{$\Delta$}\\\textbf{Baseline}\end{tabular}
    & \begin{tabular}[c]{@{}>{\itshape}c@{}}Observation-\\Conditioned\\Transitions\end{tabular}
    & \begin{tabular}[c]{@{}>{\itshape}c@{}}Physical\\Simulation\end{tabular}
    & \begin{tabular}[c]{@{}>{\itshape}c@{}}Adaptive Task\\Completion\end{tabular}
    & \begin{tabular}[c]{@{}>{\itshape}c@{}}Physical\\Properties\end{tabular}
    & \begin{tabular}[c]{@{}>{\itshape}c@{}}Visual\\Cues\end{tabular}
    & \begin{tabular}[c]{@{}>{\itshape}c@{}}In-Context\\Procedural\\Learning\end{tabular}
    & \begin{tabular}[c]{@{}>{\itshape}c@{}}Logic and\\Rules\end{tabular}
    & \begin{tabular}[c]{@{}>{\itshape}c@{}}Visual Goal\\Realization\end{tabular}
    & \textbf{Overall} \\
    \midrule
    \rule{0pt}{2.45ex}GPT-Image-2.5 & \genInterleaved~Interleaved & -- & 85.6 & 69.6 & 66.1 & 73.6 & 76.7 & 71.6 & 74.3 & 61.3 & 72.8 \\
    \rule{0pt}{2.45ex}GPT-Image-2.5 & \genGrid~Grid & Interleaved & 57.8 ($-27.8$) & 61.8 ($-7.8$) & 48.2 ($-17.9$) & 51.0 ($-22.6$) & 46.0 ($-30.7$) & 60.5 ($-11.1$) & 23.0 ($-51.3$) & 39.6 ($-21.7$) & 50.1 ($-22.7$) \\
    \rule{0pt}{2.45ex}Nano Banana Pro & \genInterleaved~Interleaved & -- & 36.2 & 56.1 & 43.5 & 49.9 & 33.7 & 43.3 & 41.4 & 28.5 & 42.8 \\
    \rule{0pt}{2.45ex}Nano Banana Pro & \genGrid~Grid & Interleaved & 39.8 ($+3.6$) & 43.2 ($-12.9$) & 42.0 ($-1.5$) & 53.1 ($+3.2$) & 42.9 ($+9.2$) & 41.4 ($-1.9$) & 31.6 ($-9.8$) & 29.9 ($+1.4$) & 40.6 ($-2.2$) \\
    \bottomrule
  \end{tabular}%
  }
\end{table}

\par\noindent\textbf{Interleaved generation outperforms single-image grids but does not fully explain performance gaps.}\quad
Interleaved generation allows the language model to guide successive visual steps, potentially reducing the reasoning burden on the renderer compared with generating all steps within a single image grid. Switching from interleaved generation to grid output reduces GPT-Image-2.5's overall score by 22.7 points, with the most obvious drop in L4 Logic \& Rules, from 74.3 to 23.0. However, output format alone does not explain the performance gap across model families: GPT-Image-2.5 with grid output still outperforms interleaved Nano Banana Pro overall, and Nano Banana Pro retains much of its performance under grid output, surpassing both Seedream 5.0 Pro and Seedream 4.5 with the same generation scheme in Table~\ref{tab:main-results}. These results suggest that interleaving is a valuable capability based on underlying visual generative performance.

\par\noindent\textbf{Interleaved generation faces the challenge of accumulated context.}\quad
While interleaving enables step-by-step guidance, it also introduces a growing context burden, challenging both the language enhancer and the renderer. To investigate this challenge, we analyze step-wise accuracy over 195 four-step cases (excluding cases that provide a direct endpoint reference), using the designated-step evaluation protocol detailed in Appendix~\prompttemplateref{app:designated-step-accuracy}{I.3}. Weaker models exhibit visible degradation: SenseNova-U1 falls from 42.6 to 27.2 (\(-15.4\)). In contrast, GPT-Image-2.5 remains relatively stable. The default Nano Banana Pro declines from 67.7 to 48.7 (\(-19.0\)). Equipping it with GPT-6 Astra reduces the first-to-last-step drop, from 82.6 to 76.4 (\(-6.2\)), although accuracy fluctuates across intermediate steps. These results suggest that maintaining reliable generation as context accumulates remains an important challenge for unified systems.

\subsubsection{Towards agentic visual generation with coding tools}
\label{sec:agentic-visual-generation}

The preceding experiments show that frontier visual generative models can increasingly extract and use visual information, enabling strong language models to apply the broad world knowledge they have acquired largely from digital worlds such as web pages and code repositories. Still, even strong systems struggle with L4-level tasks that require reasoning across successive stages of a process.

This raises a natural question: can a powerful language agent externalize its world knowledge when modeling real-world tasks through agentic inspection-generation loops and code-based tools? To investigate this question, we build two simple agentic harnesses based on Codex~\citep{openai2025codex} with the same generation budget. The first encourages the agent to perform agentic inspect-and-revise loops while prohibiting coding tools, whereas the second requires the agent to construct and inspect intermediate visual artifacts using code or modeling tools before generation, with optional refinement during generation.

\begin{figure}[H]
  \centering
  \includegraphics[width=\linewidth,pagebox=cropbox]{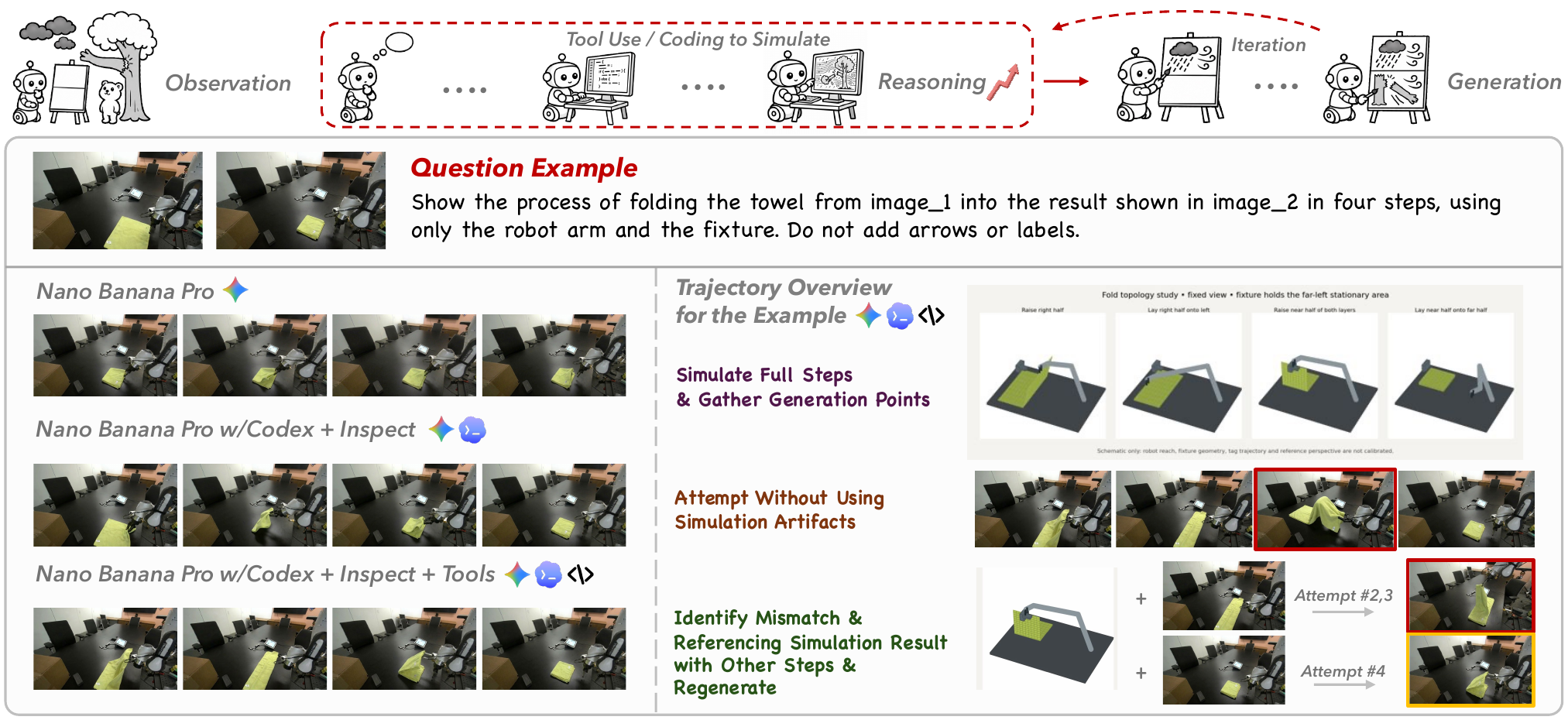}
  \caption{Visual coding as procedural guidance. Left: Result comparisons with different model harnesses. Right: code-generated
  intermediate states guide refinements of a towel-folding sequence.}
  \label{fig:agentic-comparison}
\end{figure}

\begin{table}[H]
  \centering
  \small
  \setlength{\tabcolsep}{1.8pt}
  \setlength{\abovecaptionskip}{6pt}
  \setlength{\belowcaptionskip}{5pt}
  \caption{Comparisons of harnesses for agentic visual generation using GPT-6 Astra as the language model. The coding-first inspection loop yields the best overall performance.}
  \label{tab:agentic-comparison}
  \renewcommand{\arraystretch}{1.18}
  \resizebox{\linewidth}{!}{%
  \begin{tabular}{l|c|*{8}{c}|c}
    \toprule
    \begin{tabular}[c]{@{}c@{}}\textbf{Model Paired with}\\\textbf{GPT-6 Astra}\end{tabular}
    & \begin{tabular}[c]{@{}c@{}}\textbf{Harness}\\\textbf{Enhancement}\end{tabular}
    & \begin{tabular}[c]{@{}>{\itshape}c@{}}Observation-\\Conditioned\\Transitions\end{tabular}
    & \begin{tabular}[c]{@{}>{\itshape}c@{}}Physical\\Simulation\end{tabular}
    & \begin{tabular}[c]{@{}>{\itshape}c@{}}Adaptive Task\\Completion\end{tabular}
    & \begin{tabular}[c]{@{}>{\itshape}c@{}}Physical\\Properties\end{tabular}
    & \begin{tabular}[c]{@{}>{\itshape}c@{}}Visual\\Cues\end{tabular}
    & \begin{tabular}[c]{@{}>{\itshape}c@{}}In-Context\\Procedural\\Learning\end{tabular}
    & \begin{tabular}[c]{@{}>{\itshape}c@{}}Logic and\\Rules\end{tabular}
    & \begin{tabular}[c]{@{}>{\itshape}c@{}}Visual Goal\\Realization\end{tabular}
    & \textbf{Overall} \\
    \midrule
    \multirow{3}{*}{GPT-Image-2.5} & - (GPT-6 Responses, baseline) & 85.6 & 69.6 & 66.1 & 73.6 & \textbf{76.7} & 71.6 & 74.3 & 61.3 & 72.8 \\
    & Inspection loop & \textbf{92.4} ($+6.8$) & 86.2 ($+16.6$) & 73.2 ($+7.1$) & \textbf{84.4} ($+10.8$) & 69.5 ($-7.2$) & 72.9 ($+1.3$) & 81.1 ($+6.8$) & 69.2 ($+7.9$) & 80.2 ($+7.4$) \\
    & Coding-first inspection loop & 90.7 ($+5.1$) & \textbf{88.6} ($+19.0$) & \textbf{74.9} ($+8.8$) & 80.1 ($+6.5$) & 74.5 ($-2.2$) & \textbf{74.3} ($+2.7$) & \textbf{85.7} ($+11.4$) & \textbf{81.3} ($+20.0$) & \textbf{82.4} ($+9.6$) \\
    \multirow{3}{*}{Nano Banana Pro} & - (GPT-6 Responses, baseline) & 58.9 & 68.1 & 56.7 & 64.0 & 60.1 & 63.9 & 66.9 & 48.3 & 61.2 \\
    & Inspection loop & 74.7 ($+15.8$) & 82.1 ($+14.0$) & 65.3 ($+8.6$) & 77.2 ($+13.2$) & 68.6 ($+8.5$) & 69.8 ($+5.9$) & 74.3 ($+7.4$) & 60.6 ($+12.3$) & 72.5 ($+11.3$) \\
    & Coding-first inspection loop & 78.1 ($+19.2$) & 82.0 ($+13.9$) & 65.8 ($+9.1$) & 72.1 ($+8.1$) & 63.3 ($+3.2$) & 66.7 ($+2.8$) & 81.9 ($+15.0$) & 62.3 ($+14.0$) & 72.9 ($+11.7$) \\
    \bottomrule
  \end{tabular}%
  }
\end{table}

As shown in Table~\ref{tab:agentic-comparison}, iterative agentic refinement directly raises GPT-Image-2.5 from 72.8 to 80.2 and GPT-6-enhanced Nano Banana Pro from 61.2 to 72.5. Coding tools further push the highest overall score to 82.4, but with less uniform gains across models and domains. The coding harness yields particularly large improvements on L4 Visual Goal Realization and Logic \& Rules, and L3 Physical Simulation, where visual coding makes physical or logical intermediate states explicit. Coding artifacts can remain useful even when they do not directly enter the visual generator. Figure~\ref{fig:agentic-comparison} illustrates an example trajectory, where the agent first constructs a coded simulation but decides not to use it as a direct visual reference, instead using it as prompt guidance to generate the correct first step. It then inspects the generated sequence and finds the erroneous step, uses the simulated result as a reference but obtains implausible towels, and eventually finds a solution by working backward from the final-state reference. 

While the result improves on the baselines, this trajectory reveals that incorporating complex reference information still places a considerable burden on the generator and may also help explain the uneven gains from coding, since additional guidance can sometimes fail to take effect or degrade performance. As coding agents grow more capable almost by the day, this trend presents both opportunities and challenges for future visual generative systems: they should be able to accept heterogeneous artifacts as guidance from the agent and dynamically interpret and use them, rather than treating them merely as pixel-level anchors. 

Still, code-based simulation can be computationally expensive, with a single run sometimes taking more than one hour. Distilling the code-simulating to visual-rendering process into direct visual generation capability of unified multimodal models during training could enable faster simulation inference for those complex queries.

\subsection{Validity of the Agentic Video Adapter and LMM-as-a-Judge}
\label{sec:validation}

\noindent\begin{minipage}{\linewidth}
  \centering
  \setlength{\abovecaptionskip}{6pt}
  \setlength{\belowcaptionskip}{0pt}
  \includegraphics[width=\linewidth]{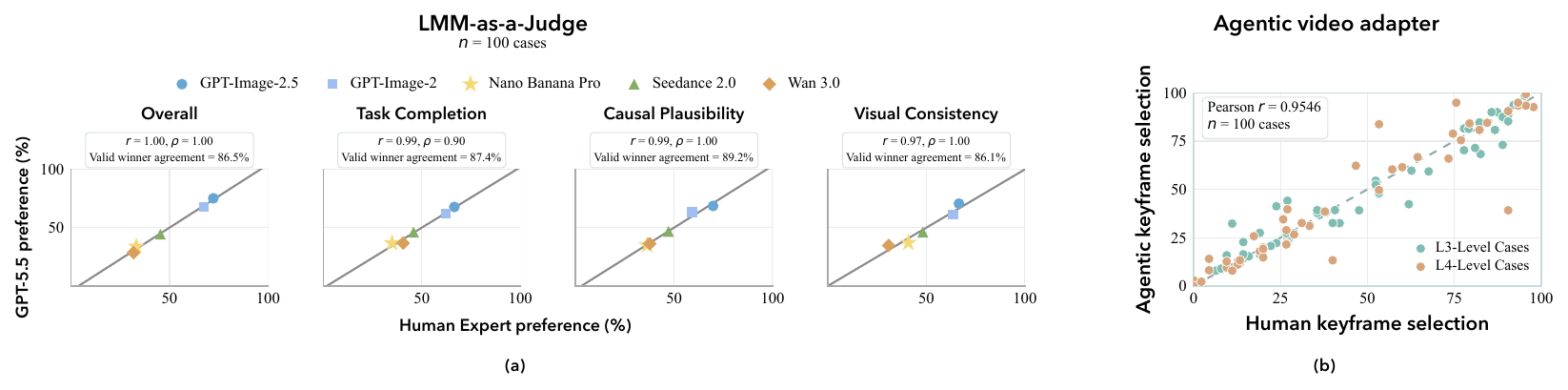}\par
    \captionof{figure}{Validation of the automatic evaluation pipeline.
  (a) Correlation between human and GPT-5.5 win rates. Scores (win = 1, tie = 0.5, loss = 0) are averaged across pairwise comparisons for each model.
  (b) Correlation between evaluator scores from agent-selected and human-selected keyframes.}
  \label{fig:validation-correlations}
\end{minipage}\par

We validate both \emph{LMM-as-a-Judge} and \emph{Agentic video adapter}. For the \emph{LMM-as-a-Judge}, following VBench~\citep{huang2024vbench}, six human experts perform blind pairwise comparisons among five representative systems. As shown in Figure~\ref{fig:validation-correlations}(a), \benchmark{} and human win rates achieve Pearson $r\geq0.97$ across four dimensions (Overall, TC, CP, and VC), with winner agreement rates $\geq86\%$ among valid human-majority winner votes. For the \emph{agentic video adapter}, three annotators select keyframes for 100 videos generated on our benchmark. We score human- and adapter-selected keyframes using the same evaluator. The resulting scores achieve a Pearson correlation of $0.9546$ (Figure~\ref{fig:validation-correlations}(b)), supporting the reliability of the automated protocol.

\section{Conclusion}

In this work, we present \benchmark{}, a benchmark for exploring world state prediction as an emerging capability of visual generative models. \benchmark{} evaluates L3--L4 capabilities based on a cognitively motivated four-level hierarchy, through 320 manually designed cases across eight categories. Experiments on image and video systems with a shared evaluation interface show that frontier unified models exhibit early-stage world state prediction capabilities on real-world tasks, while substantial gaps remain across models and become more pronounced when tasks require heterogeneous context and tightly coupled reasoning and generation. Our analyses further identify potential performance gains from interleaved generation, stronger language-model reasoning, and iterative agentic refinement for this task, while highlighting challenges that emerge as models advance in these directions. We hope that \benchmark{} will serve as a useful testbed for developing unified generative models that can integrate heterogeneous context, agentic reasoning and visual generation to predict coherent world evolution.

\clearpage
\bibliography{references}
\bibliographystyle{assets/plainnat}

\clearpage
\appendix
\section*{Appendix}

\section{Related Work}

\paragraph{World Modeling.}
World models predict how an environment evolves under actions. Genie learns action-controllable video dynamics~\citep{bruce2024genie}, and GameNGen demonstrates real-time game simulation~\citep{valevski2024gamen}, both emphasizing temporal continuity. Video-based world-model benchmarks also assess scene-conditioned reactivity and multi-turn interaction~\citep{yang2026worldexam,ying2026wbench}. \benchmark{} instead focuses on key visual states to evaluate the intelligence of generative systems. Although these world models support interactive rollouts, they do not currently support the reasoning-based transitions and heterogeneous references required by our L3-L4 tasks and are thus excluded from our benchmark.

\paragraph{Unified Multimodal Systems.}
Unified multimodal models (UMMs) bring understanding and generation into a shared
interface. Chameleon explores mixed-modal early fusion~\citep{chameleon2024},
while Show-o~\citep{xie2024showo} and Janus~\citep{wu2024janus} develop unified
architectures for understanding and generation. BAGEL~\citep{deng2025bagel},
Emu3.5~\citep{cui2025emu35}, and SenseNova-U1~\citep{diao2026sensenovau1}
further support interleaved multimodal inputs and outputs. In a broader sense, generative systems that connect language-side reasoning with visual generation through a shared interface can also be considered unified even without a unified architecture, for example, a dedicated prompt enhancer paired with an image or video generation model~\citep{seedream2025seedream40,wanPE}. \benchmark{} evaluates model performance regardless of the underlying unification approach.

\paragraph{Multimodal References for Generation.}
Visual references can guide appearance, identity, and spatial structure~\citep{ye2023ipadapter,wang2024msdiffusion}, as well as motion~\citep{yan2026scail2,wang2024motionctrl}. Visual drafts, including those rendered from executable code, also provide intermediate guidance for generation~\citep{jiang2025draco,li2026coco,li2026videococo}. \benchmark{} focuses on interpreting heterogeneous references---including prior procedural experience, abstract goals, physical properties, and visual rules---to predict successive world states. Our experiments further investigate whether coding agents can construct procedural examples and intermediate visual states that turn such information into useful guidance for visual generation.

\paragraph{Evaluation of Unified Systems.}
Existing evaluations increasingly move beyond fidelity and shallow
text--image alignment. WISE evaluates world-knowledge-informed text-to-image
semantics~\citep{wise}, GENIUS studies pattern induction and constraint
execution~\citep{genius}, UEval evaluates interleaved image--text
responses~\citep{li2026ueval}, and MICON-Bench and MMIG-Bench study multi-image
context~\citep{wu2026micon,hua2025mmig}. VIBE evaluates visual instruction following in image editing~\citep{zhang2026vibe}. GIR-Bench~\citep{li2026girbench} uses understanding and reasoning to guide generation,
whereas \benchmark{} evaluates more tightly coupled understanding and generation:
the same image may simultaneously specify visual content and encode procedural
or logical constraints, requiring models to interpret and apply both jointly. Some tasks in those benchmarks approach L3 reasoning in our hierarchy, but primarily assess a static visual result.

\section{Task Formulation}
\label{app:formulation}

\benchmark{} formulates context-grounded visual world prediction as generating
an ordered sequence of visual states from interleaved image--text context. Each
test case consists of a natural-language instruction $p$, one or more input
images $\mathcal{X}=(x_m)_{m=1}^{M}$, and an ordered sequence of target output
images $\mathcal{Y}=(y_k)_{k=1}^{K}$. The input images may serve different
functions---for example, specifying an initial state, a goal, an object or scene
reference, a physical property, a visual rule, or an intermediate
demonstration. Given the multimodal context $(p,\mathcal{X})$, a generative
model $G$ is required to produce
\begin{equation}
  \widehat{\mathcal{Y}} = G(p,\mathcal{X})
  = (\hat{y}_1,\hat{y}_2,\ldots,\hat{y}_K),
\end{equation}
where the generated images must jointly satisfy the instruction and visual
constraints while forming a temporally ordered, causally coherent process.
This multimodal-context-to-visual-sequence formulation provides a common
interface for different model families: an image model may generate the
sequence jointly or through interleaved generation, whereas a video model may produce a continuous
clip whose selected frames constitute $\widehat{\mathcal{Y}}$.

\section{Data Collection}
\label{app:data-collection}

We combine real-world datasets, online videos, illustrated manuals, gameplay footage,
and manually designed scenarios. When suitable visual material is unavailable,
we synthesize or enhance analogous cases with image-generation
tools~\citep{google2025nanobanana,longcat2025image,openai2025gptimage} while
preserving the intended spatial, physical, and procedural constraints. The
resulting output-step distribution is summarized in
Figure~\ref{fig:data-collection-distribution}. The category-specific collection
procedures are as follows:

\begin{center}
  \includegraphics[width=0.60\linewidth]{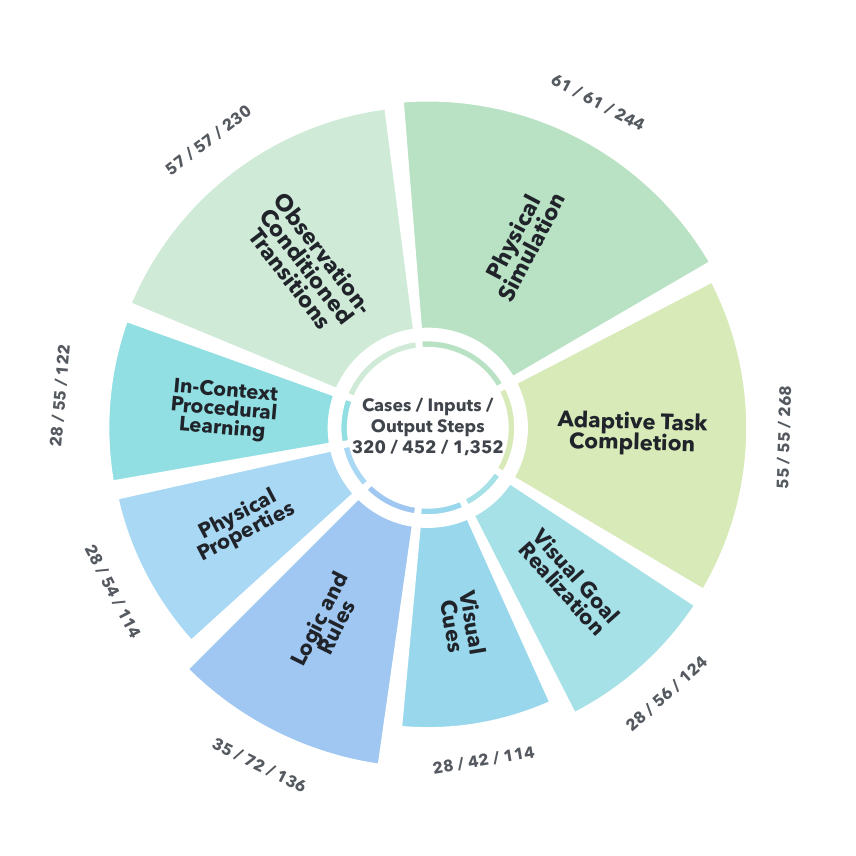}
  \captionof{figure}{Data distribution across the eight \benchmark{}
  categories. Labels report cases/input images/requested visual states.}
  \label{fig:data-collection-distribution}
\end{center}

\begingroup
\setlength{\parskip}{2pt}
\noindent\textbullet\ \textbf{Observation-Conditioned Transitions.} We collect cases from
  egocentric and game data~\citep{damen2018epickitchens,khazatsky2024droid}.
  We retain scenarios in which the subsequent action logic must be grounded in
  observations from the initial frame.
\par\noindent\textbullet\ \textbf{Physical Simulation.} We source physical events from game and
  egocentric data, science and educational videos, and filtered Internet
  examples. We further synthesize analogous scenarios based on real-world
  experience, retaining cases whose evolution requires reasoning from the
  observed initial state.
\par\noindent\textbullet\ \textbf{Adaptive Task Completion.} Most cases are collected from game
  and egocentric data; additional everyday tasks are designed and synthesized
  from common real-life experience.
\par\noindent\textbullet\ \textbf{Visual Cues.} We construct cues by drawing or manually marking
  images and by annotating collected photographs. Cases may contain either a
  single reference or multiple references. For multi-view cases, we collect
  home-design views and use Claude Opus~\citep{anthropic2025claudeopus4} to operate Blender to create
  geometrically matched source images, which are then enhanced with Nano Banana
  Pro for greater realism while preserving their underlying 3D consistency.
\par\noindent\textbullet\ \textbf{Physical Properties.} We select examples from
  science videos and targeted Internet searches, and synthesize additional
  cases grounded in physical phenomena encountered in everyday experience.
\par\noindent\textbullet\ \textbf{Visual Goal Realization.} We extract initial and
  target states from assembly and building videos on YouTube, occasionally
  enhancing the target image for clarity and visual quality.
\par\noindent\textbullet\ \textbf{Logic and Rules.} We draw cases from existing games and use
  Claude Opus~\citep{anthropic2025claudeopus4} to synthesize additional games with controlled visual rules.
\par\noindent\textbullet\ \textbf{In-Context Procedural Learning.} We collect illustrated
  manuals and pair their procedures with corresponding or analogous initial
  states, including cases that require transferring a procedure across
  visually different situations.
\par\endgroup

\section{Evaluation Details}
\label{app:evaluation-details}
\label{app:evaluation-rubrics}

\subsection{Details on Human Hints}
\label{app:human-hints}

Each test case includes a set of human-written hints for evaluation. These
hints are provided only to the judge and are never exposed to the generation
model. They reduce ambiguity in otherwise underspecified instructions and draw
the judge's attention to task-critical evidence that may appear only at a
particular stage of the generated sequence.

Annotators write the hints after jointly inspecting the instruction, all
multimodal references, and the expected progression. A hint may identify
task-critical objects and their reference bindings, required intermediate
milestones and temporal ordering, or physical and logical constraints that
must remain valid across states. When a task admits multiple valid procedures,
the hints specify the indispensable outcome or constraint without prescribing
unnecessary appearance details. The same case-specific hints are used to
evaluate every model output. They guide the assessment of PA, OC, CP, and VC, but
do not substitute for visible evidence in the generated sequence.
The same Task Completion (TC) hints are provided to both Progress Alignment
(PA) and Outcome Completion (OC) evaluations.
Figure~\ref{fig:detailed-qualitative-examples} provides examples of these
human-annotated hints.

\subsection{Evaluation Procedure}

For every case, the judge is given the natural-language instruction, any
judge-only hints, all reference and initial images, and the complete ordered
sequence of generated images. The judge first reads the full sequence, evaluates
the relevant local transitions, and then performs a global pass to resolve
dependencies that may not be apparent from an isolated frame. Credit is based
on visible evidence; captions, arrows, or a plausible narrative do not by
themselves establish that a state change occurred.

\subsection{Task Completion}
\label{app:task-completion}

We compute TC from Progress Alignment (PA) and Outcome Completion (OC). PA
evaluates the complete ordered sequence on a three-level scale (0-2). A
correct-looking endpoint does not compensate for a missing critical step.
OC evaluates the final generated state. It receives a binary signal whether every indispensable endpoint requirement is visibly satisfied. After normalizing
PA to $[0,1]$, the two components are combined using a case-specific outcome
weight $w_o\in[0,0.5]$:
\begin{equation}
  \mathrm{TC}=(1-w_o)\,\mathrm{PA}+w_o\,\mathrm{OC}.
\end{equation}
The outcome contributes at most 50\% of TC. Its weight is set per case
according to the importance of the requested result. Process-focused cases
use $w_o=0$: OC is not evaluated separately, and TC
depends solely on PA. Detailed scoring criteria are provided in the
\hyperref[app:pa-prompt]{PA} and \hyperref[app:oc-prompt]{OC evaluator templates}
in Appendix~\prompttemplateref{app:evaluator-prompts}{I.3}.

\subsection{Causal Plausibility}

CP evaluates whether the complete sequence follows the applicable physical,
logical, game, or task-specific rules on a four-level scale (0-3).
The judge checks adjacent cause--effect relations as well as long-range
dependencies, object conservation, prerequisites, contact and support,
topology, temporal order, and task-specific constraints. Detailed scoring criteria are provided
in the \hyperref[app:cp-prompt]{CP evaluator template} in
Appendix~\prompttemplateref{app:cp-prompt}{I.3}.

\subsection{Visual Consistency}

VC evaluates whether task-relevant identities and scene attributes remain
consistent across the sequence on a four-level scale (0-3). It checks for unexplained changes in object
identity, count, geometry, scale, material, camera, and background, including
disappearance, duplication, deformation, teleportation, and scene resets.
The judge compares both adjacent and non-adjacent frames to capture cumulative
drift, while exempting changes required by the task or visibly established by
the causal process. Detailed scoring criteria are provided in the
\hyperref[app:vc-prompt]{VC evaluator template} in
Appendix~\prompttemplateref{app:vc-prompt}{I.3}.

\subsection{Score Aggregation}

We first express the scores derived from discrete ratings on a common 0--3
scale: TC from Section~\ref{app:task-completion} is multiplied by 3, while CP
and VC retain their original 0--3 scores. We then compute their weighted
aggregate on this scale and convert it to a 100-point score:
\begin{equation}
  \mathrm{Final} = \frac{100}{3}\left(\alpha_1\,\mathrm{TC} + \alpha_2\,\mathrm{CP}
  + \alpha_3\,\mathrm{VC}\right).
\end{equation}
We set $\alpha_1=\alpha_2=0.4$ and $\alpha_3=0.2$: TC and CP receive equal weight, while VC receives half the weight of either.
These aggregation weights are fixed across categories and are distinct from
the case-specific outcome weight $w_o$ used within TC. We report both the
individual metric scores and their weighted aggregate on the 100-point scale.

\subsection{Performance Across Evaluation Dimensions}
\label{app:metric-comparison}
Figure~\ref{fig:metric-comparison} shows how representative models' scores are distributed on \benchmark{}.

\noindent\begin{minipage}{\linewidth}
  \centering
  \includegraphics[width=0.40\linewidth]{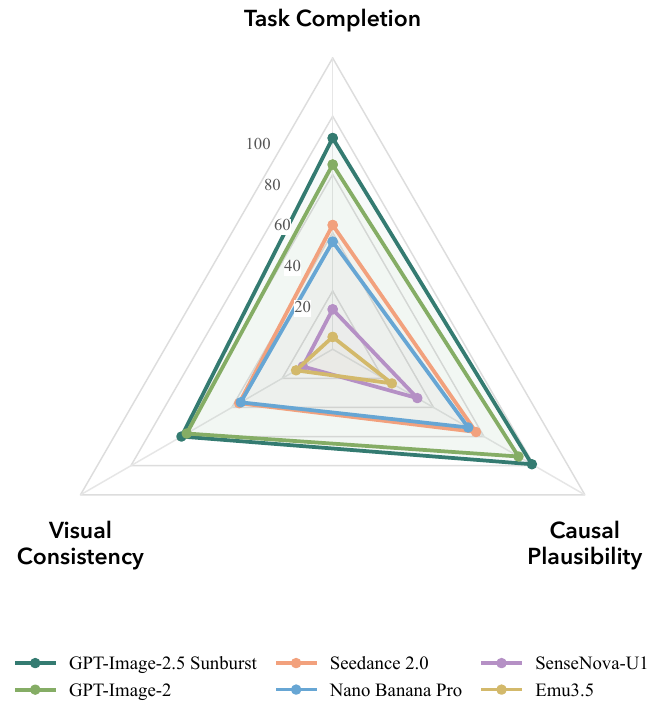}\par
  \setlength{\abovecaptionskip}{6pt}
  \captionof{figure}{Performance across TC, CP, and VC on \benchmark{}.}
  \label{fig:metric-comparison}
\end{minipage}

\subsection{Parameter Settings}
\label{app:parameter-settings}

We use GPT-5.5 as the judge with reasoning effort set to \texttt{medium}.
All GPT-6 configurations use GPT-6 Astra with \texttt{medium} reasoning effort,
whether accessed through the Responses API or used as an agent through Codex.
Our Codex-based experiments use \texttt{codex-cli 0.153.4}.

\clearpage
\section{Stability of LMM-as-a-Judge}
\label{app:model-stability}

To assess sensitivity to evaluator choice, we rescore the same model outputs
with the open-source GLM-5.3-Flash~\citep{zai2026glm53flash},
GPT-5.5~\citep{openai2026gpt5.5}, and Gemini 3.1
Pro~\citep{google2026gemini31pro}. Figure~\ref{fig:model-stability} shows a
systematic calibration difference: GLM-5.3-Flash assigns higher absolute
scores, whereas Gemini 3.1 Pro is more conservative. Nevertheless, the model
ordering is largely preserved across all three judges, suggesting that our
relative conclusions are robust to evaluator choice even when absolute score
scales differ.

\noindent\begin{minipage}{\linewidth}
  \centering
  \includegraphics[width=0.97\linewidth]{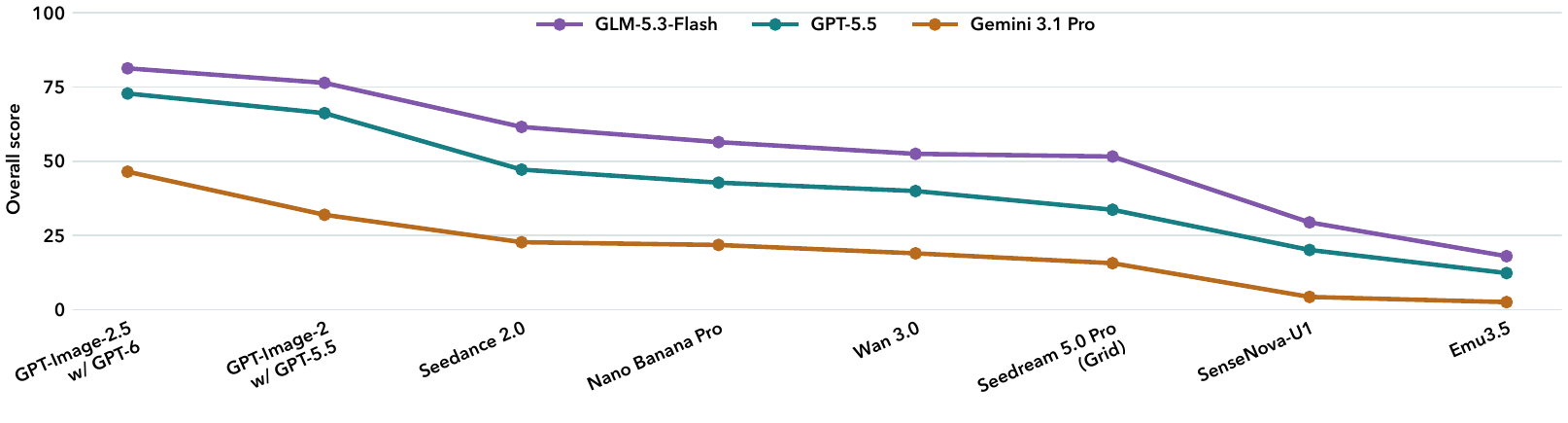}
  \captionsetup{justification=centering}
  \captionof{figure}{Overall model scores assigned by three LMM judges.}
  \label{fig:model-stability}
\end{minipage}\par

\paragraph{Model stability across repeated generations.}
We further repeat generation for three rounds and score every run with GPT-5.5.
Table~\ref{tab:judge-variance} reports the resulting mean and standard deviation
for models with complete repeated runs. Means are computed over all 320 cases,
and standard deviations are computed across generation-run means. Overall-score
standard deviations do not exceed 0.53, although individual categories show
greater variability, reaching 2.7 points.

\begingroup
\newcommand{\mstd}[2]{#1\,{\fontsize{9.5}{11}\selectfont$\pm$\,\num[round-mode=places,round-precision=2,round-pad=true]{#2}}}
\begin{center}
  \centering
  \small
  \setlength{\tabcolsep}{2.8pt}
  \renewcommand{\arraystretch}{1.5}
  \resizebox{0.97\linewidth}{!}{%
  \begin{tabular}{>{\fontsize{10}{12}\selectfont\raggedright\arraybackslash}m{3.0cm}|
                  *{3}{>{\fontsize{11}{13}\selectfont\centering\arraybackslash}m{2.0cm}}|
                  *{5}{>{\fontsize{11}{13}\selectfont\centering\arraybackslash}m{2.0cm}}|
                  >{\fontsize{11}{13}\selectfont\centering\arraybackslash}m{2.0cm}}
    \specialrule{0.8pt}{0pt}{0pt}
    \multirow{2}{*}[-8pt]{\fontsize{11}{13}\selectfont\textbf{Method}}
    & \multicolumn{3}{c|}{\fontsize{11}{13}\selectfont\textbf{L3 Tasks}}
    & \multicolumn{5}{c|}{\fontsize{11}{13}\selectfont\textbf{L4 Tasks}}
    & \multirow{2}{*}[-8pt]{\fontsize{11}{13}\selectfont\textbf{Overall}} \\
    \cline{2-9}
    & \fontsize{9.5}{11}\selectfont\itshape\shortstack[c]{Observation-\\Conditioned\\Transitions}
    & \fontsize{9.5}{11}\selectfont\itshape\shortstack[c]{Physical\\Simulation}
    & \fontsize{9.5}{11}\selectfont\itshape\shortstack[c]{Adaptive Task\\Completion}
    & \fontsize{9.5}{11}\selectfont\itshape\shortstack[c]{Physical\\Properties}
    & \fontsize{9.5}{11}\selectfont\itshape\shortstack[c]{Visual\\Cues}
    & \fontsize{9.5}{11}\selectfont\itshape\shortstack[c]{In-Context\\Procedural\\Learning}
    & \fontsize{9.5}{11}\selectfont\itshape\shortstack[c]{Logic and\\Rules}
    & \fontsize{9.5}{11}\selectfont\itshape\shortstack[c]{Visual Goal\\Realization}
    & \\
    \specialrule{0.5pt}{0pt}{0pt}
    SenseNova-U1
      & \mstd{24.1}{0.73} & \mstd{24.6}{0.43} & \mstd{24.7}{0.39}
      & \mstd{19.9}{0.84} & \mstd{19.1}{1.21} & \mstd{20.1}{0.45} & \mstd{5.9}{0.34} & \mstd{12.2}{0.21} & \mstd{20.1}{0.30} \\
    Emu3.5
      & \mstd{17.4}{0.43} & \mstd{13.8}{0.29} & \mstd{18.5}{0.84}
      & \mstd{6.6}{0.49} & \mstd{10.3}{0.71} & \mstd{8.1}{0.54} & \mstd{4.8}{0.50} & \mstd{7.6}{0.56} & \mstd{12.3}{0.16} \\
    BAGEL
      & \mstd{4.9}{0.43} & \mstd{7.0}{0.42} & \mstd{8.3}{0.28}
      & \mstd{8.5}{0.48} & \mstd{12.5}{0.11} & \mstd{7.0}{1.07} & \mstd{2.9}{0.32} & \mstd{4.0}{0.67} & \mstd{6.8}{0.13} \\
    Qwen-Image
      & \mstd{7.8}{0.29} & \mstd{6.8}{0.16} & \mstd{6.2}{0.63}
      & \mstd{4.0}{0.56} & \mstd{6.9}{0.31} & \mstd{6.1}{0.49} & \mstd{0.6}{0.16} & \mstd{3.4}{0.21} & \mstd{5.6}{0.12} \\
    \hline
    GPT-Image-2.5
      & \mstd{85.6}{1.12} & \mstd{69.6}{0.79} & \mstd{66.1}{1.01}
      & \mstd{73.6}{1.84} & \mstd{76.7}{1.08} & \mstd{71.6}{0.92} & \mstd{74.3}{1.13} & \mstd{61.3}{0.74} & \mstd{72.8}{0.39} \\
    GPT-Image-2
      & \mstd{74.2}{0.74} & \mstd{70.6}{1.12} & \mstd{59.6}{0.95}
      & \mstd{66.0}{0.72} & \mstd{62.6}{0.70} & \mstd{70.6}{0.76} & \mstd{66.6}{0.16} & \mstd{51.8}{2.68} & \mstd{66.2}{0.53} \\
    Nano Banana Pro
      & \mstd{36.2}{1.15} & \mstd{56.1}{0.95} & \mstd{43.5}{0.40}
      & \mstd{49.9}{0.42} & \mstd{33.7}{0.62} & \mstd{43.3}{1.55} & \mstd{41.4}{0.81} & \mstd{28.5}{0.50} & \mstd{42.8}{0.48} \\
    Seedream 5.0 Pro
      & \mstd{30.3}{2.00} & \mstd{45.9}{0.69} & \mstd{33.4}{1.02}
      & \mstd{46.7}{1.40} & \mstd{27.7}{1.16} & \mstd{38.9}{0.76} & \mstd{11.9}{0.48} & \mstd{29.1}{0.63} & \mstd{33.6}{0.19} \\
    Seedream 4.5
      & \mstd{19.4}{0.79} & \mstd{14.9}{0.13} & \mstd{18.7}{0.14}
      & \mstd{19.7}{1.14} & \mstd{16.0}{0.49} & \mstd{21.4}{1.03} & \mstd{3.4}{0.38} & \mstd{12.8}{0.71} & \mstd{16.0}{0.03} \\
    \hline
    Seedance 2.0
      & \mstd{53.3}{0.86} & \mstd{53.4}{0.83} & \mstd{48.7}{0.46}
      & \mstd{40.8}{1.32} & \mstd{56.6}{0.79} & \mstd{55.4}{1.96} & \mstd{17.7}{0.97} & \mstd{43.3}{0.39} & \mstd{47.2}{0.08} \\
    Gemini Omni
      & \mstd{45.6}{0.58} & \mstd{50.2}{0.26} & \mstd{41.0}{0.23}
      & \mstd{47.2}{2.49} & \mstd{42.7}{0.88} & \mstd{49.9}{1.25} & \mstd{16.2}{0.56} & \mstd{33.0}{1.76} & \mstd{41.6}{0.12} \\
    Wan 3.0
      & \mstd{51.9}{1.22} & \mstd{47.3}{2.20} & \mstd{38.5}{1.20}
      & \mstd{42.7}{1.19} & \mstd{42.7}{1.51} & \mstd{35.5}{0.98} & \mstd{17.2}{1.06} & \mstd{29.6}{0.85} & \mstd{39.9}{0.35} \\
    \specialrule{0.8pt}{0pt}{0pt}
  \end{tabular}%
  }
  \captionof{table}{Model stability across repeated generations. Scores are
  measured by GPT-5.5 and reported as mean $\pm$ standard deviation over three
  generation runs. Model names and ordering follow Table~\ref{tab:main-results}.}
  \label{tab:judge-variance}
\end{center}
\endgroup

\section{Additional Analysis on Coding Artifacts}
\label{app:coding-artifact-analysis}

The main paper (Section~\ref{sec:agentic-visual-generation}) discusses indirect uses of coding artifacts and errors introduced by direct visual conditioning. As Figure~\ref{fig:coding-artifact-analysis} shows, in most cases, the agent chooses not to pass these artifacts to the generator, but uses them to guide revisions or translates them into more precise descriptions. However, some of the largest gains occur in tasks where the agent can explicitly represent or guide the process: with the coding harness, GPT-Image-2.5 improves by 20.0 points on Visual Goal Realization and 19.0 points on Physical Simulation over the GPT-6 Responses baseline (Table~\ref{tab:agentic-comparison}). In this appendix, we therefore focus on how the agent operates when visual artifacts are used as direct inputs to the generator.

Direct use is most frequent in Logic and Rules (22 of 35 cases) and Physical Simulation (24 of 61 cases). These tasks lend themselves to externalizing otherwise latent reasoning as discrete state trajectories, geometric constructions, or physical-process sketches. The examples in Figure~\ref{fig:coding-artifact-analysis} illustrate both patterns: an overlay communicates a physically grounded progression to GPT-Image-2.5, whereas a coded state simulator supplies intermediate configurations that Nano Banana Pro renders into the requested sequence.

These cases highlight how capable current agents can be on tasks they can simulate through code. Previous work emphasizes composition through structural drafts in CoCo~\citep{li2026coco}, programmable visual sketches in GenClaw~\citep{ye2026genclaw}, and context acquisition through search in Qwen-Image-Agent~\citep{zhang2026qwenimageagent}. Our simple harness on \benchmark{}, meanwhile, demonstrates the strength of LLMs in simulating processes. Their world knowledge also enables guidance to be adapted dynamically to the task. In the Physical Simulation example, the agent recognizes that the original reference alone provides insufficiently fine-grained guidance for subtle changes in piston height, and therefore introduces guide lines and labels to specify those changes. This suggests that future enhancers should extend well beyond language enhancement to intelligently provide task-appropriate multimodal information, and the underlying visual generators must evolve alongside these advances to effectively interpret and use such guidance.

\begin{center}
  \includegraphics[width=\linewidth]{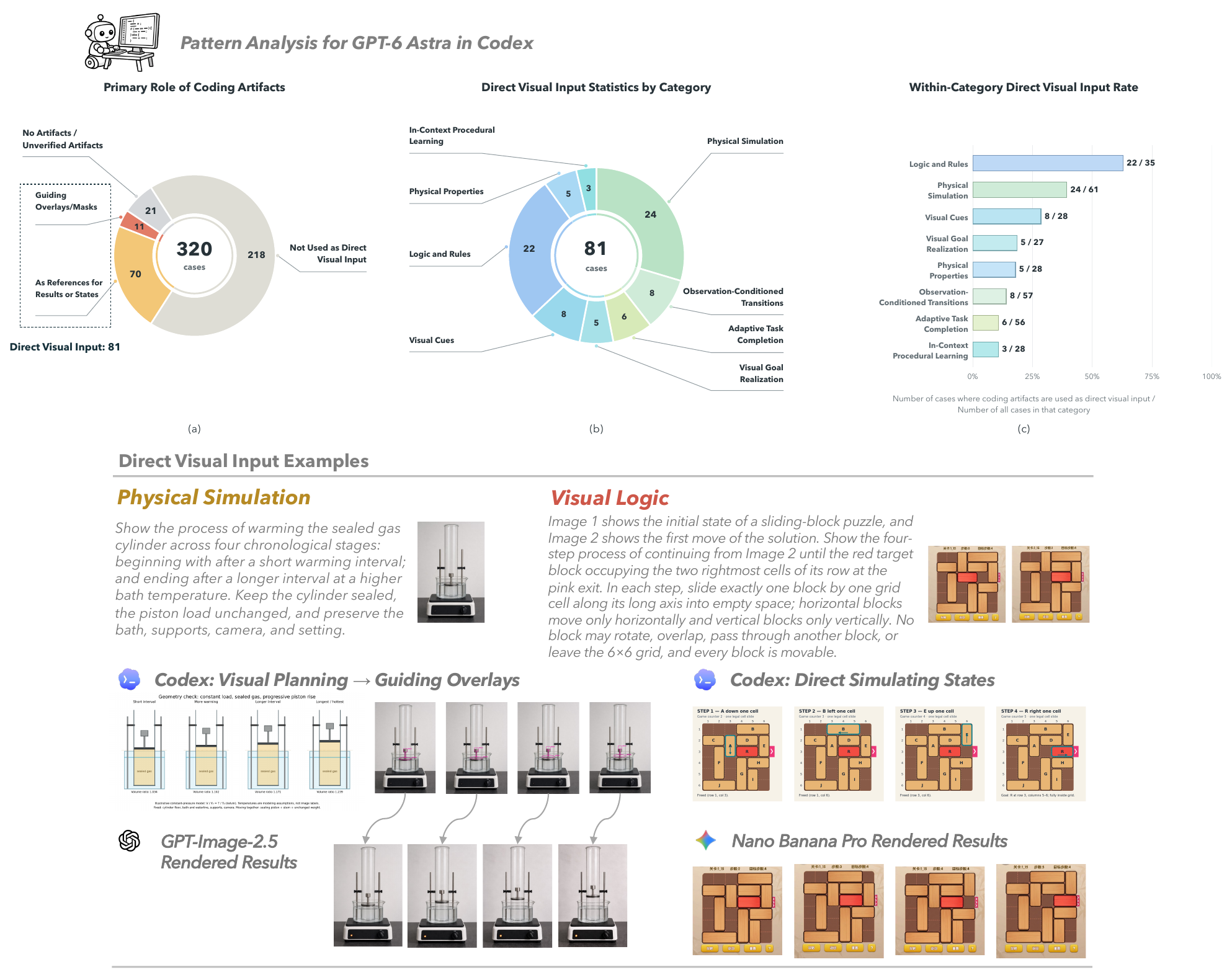}
  \captionof{figure}{Usage patterns of coding artifacts under GPT-6 Astra with
  Codex. Top: the role and category distribution of coding artifacts across 320
  cases. Bottom: representative examples using guiding overlays and simulated
  intermediate states as direct visual inputs.}
  \label{fig:coding-artifact-analysis}
\end{center}

\clearpage
\section{Detailed Qualitative Examples}
\begin{center}
  \includegraphics[width=\linewidth,height=0.89\textheight,keepaspectratio,page=1]{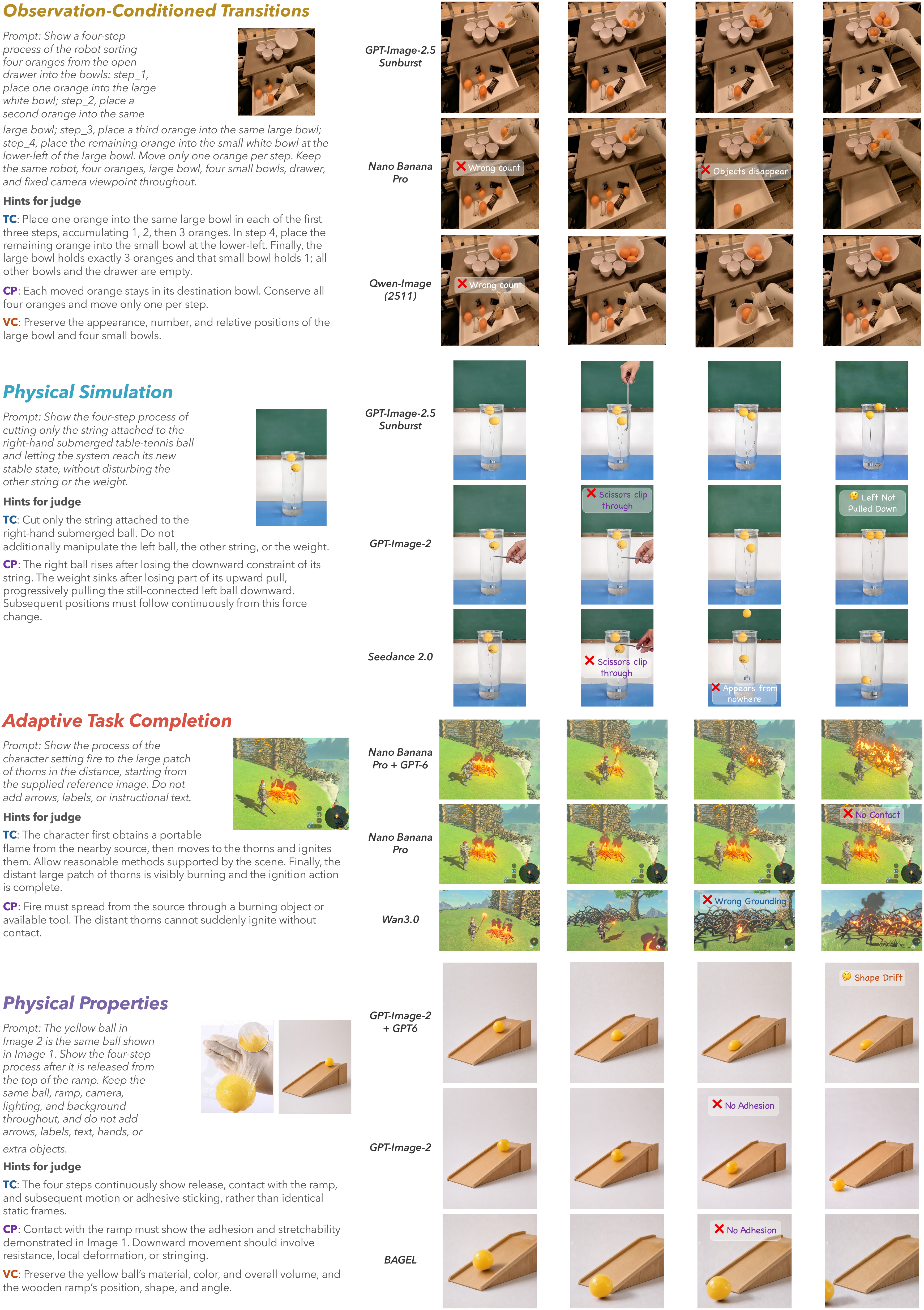}
  \captionsetup{justification=centering}
  \captionof{figure}{Detailed Qualitative Examples, Model Outputs and Error Patterns.}
  \label{fig:detailed-qualitative-examples}
\end{center}

\clearpage
\begin{center}
  \includegraphics[width=\linewidth,height=0.97\textheight,keepaspectratio,page=2]{figures/detailed_qualitative_examples.pdf}
\end{center}

\clearpage
\section{Details on the Language Model Enhancer}
\label{app:nbp-gpt6}
GPT-Image models are paired with their language models through the native image generation tool of the Responses API. For Nano Banana Pro in Section~\ref{sec:language-enhancer}, we mock this tool so that GPT-6 calls Nano Banana Pro through the Responses API, as described in Algorithm~\ref{alg:loop}. GPT-6 Astra serves as the controller $\mathcal{M}$ and Nano Banana Pro as the image generator $\mathcal{G}$: for each state, the controller writes the generation prompt and selects an ordered set of reference images from the pool, and the generator renders the state in a stateless call; each generated image is then added to the pool and returned to the controller as visual input.

\begin{algorithm}[H]
\caption{Sequential generation with GPT-6 and Nano Banana Pro}
\label{alg:loop}
\begin{algorithmic}[1]
\Require Task prompt $P$, input images $\mathcal{S}=\{s_1,\dots,s_m\}$, number of states $N$,
         controller $\mathcal{M}$, image generator $\mathcal{G}$
\Ensure Ordered image sequence $(x_1,\dots,x_N)$
\State $\mathcal{I} \gets \mathcal{S}$ \Comment{image pool visible to $\mathcal{M}$}
\State $H \gets [\,P, \mathcal{S}\,]$ \Comment{controller context}
\For{$t = 1$ \textbf{to} $N$} \Comment{one generation call per state, no retries}
    \State $(p_t, R_t) \gets \mathcal{M}(H)$
        \Comment{prompt and ordered references $R_t \subseteq \mathcal{I}$}
    \State $x_t \gets \mathcal{G}(p_t, R_t)$
        \Comment{stateless call}
    \State $\mathcal{I} \gets \mathcal{I} \cup \{x_t\}$
    \State $H \gets H \oplus [\,(p_t, R_t),\ x_t\,]$
        \Comment{$x_t$ is returned as visual input}
\EndFor
\State \Return $(x_1,\dots,x_N)$
\end{algorithmic}
\end{algorithm}

\ifrenderprompttemplates
  \section{Prompt Templates}
\label{app:prompt-templates}
\newcommand{\prompttitle}[1]{\par\medskip\noindent\textbf{#1}\par\smallskip}

\subsection{Generation-Interface Adapters}

\prompttitle{Grid generation.}
\begin{lstlisting}[style=prompt]
{{ORIGINAL_TASK}}
Output N steps, and arrange the N steps in a {{LAYOUT}} grid. Please collage all the steps into one image. Do not render any step labels, captions, instructions, explanatory text, arrows, or guide graphics inside the image unless visible text is explicitly required by the user's task.
\end{lstlisting}
Here, \texttt{\{\{LAYOUT\}\}} is one of $1\times2$, $2\times2$, or $2\times3$.

\prompttitle{Interleaved generation.}
\begin{lstlisting}[style=prompt]
{{ORIGINAL_TASK}}

Output each step with one textual instruction and one image. Don't output all steps in one image, instead output N images interleaved with text description. Do not render the textual instructions, step labels, captions, explanatory text, arrows, or guide graphics inside the images unless visible text is explicitly required by the user's task.
\end{lstlisting}

\prompttitle{Video generation.}
\begin{lstlisting}[style=prompt]
{{ORIGINAL_TASK}}

Create a single coherent video that visually carries out the requested process from beginning to end. Show every requested stage in the specified chronological order, with each stage clearly visible and connected through natural motion. Do not use a split screen or other instructional overlays.
\end{lstlisting}

\subsection{Agentic Video Adapter}
\label{app:video-adapter-prompt}

\begin{lstlisting}[style=prompt]
You are selecting semantic benchmark frames from a generated video.

Inspect the source MP4 itself. You may repeatedly use ffmpeg to sample any timestamps and compare frames. Use the text instruction, reference images, complete video, and neighboring video states to identify the {{STEP_COUNT}} frames that most clearly demonstrate the requested key steps or milestones. The selected timestamps must be strictly increasing. Do not select evenly spaced timestamps merely for convenience. If a requested state never appears, select the closest visible state in its correct chronological region and mark it as achieved=false. This is frame selection, not quality scoring.

You MUST always select and extract exactly {{STEP_COUNT}} frames, even when the video is poor, incomplete, ambiguous, or does not visibly achieve one or more requested states. In that extreme case, you may choose the temporal point based on when the milestone is expected to occur as specified in the prompt. Never omit a step, refuse selection, stop early, or leave an output file unwritten because no ideal frame exists. For every missing state, choose the semantically closest visible frame from the most appropriate chronological region; use neighboring frames and the requested order to make the best approximation. Set achieved=false and briefly explain the mismatch in evidence, but still extract that frame and include it in the JSON result.

Use this ffmpeg executable:
{{FFMPEG_EXECUTABLE}}

Extract the final full-resolution frames to exactly these paths:
{{OUTPUT_FRAME_PATHS}}

Return the matching JSON only after every file has been written. Each filename must be the basename only and exactly step_1.jpg through step_{{STEP_COUNT}}.jpg; do not return a temporary sampling path such as frames/t2.4.jpg. Before returning, verify that all {{STEP_COUNT}} required JPEG files exist, are non-empty, and correspond one-to-one with the JSON entries.

Evaluation input:
{{EVALUATION_INPUT_JSON}}

## Output Contract

{
  "steps": [
    {
      "step": 1,
      "timestamp_seconds": 0.0,
      "filename": "step_1.jpg",
      "achieved": true,
      "evidence": "Visible evidence for the requested milestone, or a brief mismatch explanation."
    }
  ],
  "overall_notes": "Brief notes about the selection."
}
\end{lstlisting}

\subsection{LMM-as-a-Judge Templates}
\label{app:evaluator-prompts}

\prompttitle{TC Sub-Metric: Progress Alignment (PA)}
\phantomsection
\label{app:pa-prompt}
\begin{lstlisting}[style=prompt]
# Progress Alignment Evaluation

Your task is to act as a PROGRESS ALIGNMENT AUDITOR in Strict Evaluation Mode. Inspect the complete ordered image sequence and judge whether the overall task unfolds in accordance with the user's intended process. Use the text instruction, HINTS, and reference images to determine the task-critical milestones, then judge whether the sequence visibly realizes those milestones in the required or logically necessary order and develops in the way the user expected. You should assess whether the whole sequence unfolds correctly and give a sequence-level evaluation; a correct final result alone is insufficient.

## Scoring Hierarchy

SCORE 2 [INTENDED PROCESS AND ALL MILESTONES REALIZED]

- The sequence's overall task unfolds in the way the user intended, and every required milestone is visibly realized rather than merely approached, initiated, implied, or described.
- The process shown across the sequence matches the expected interpretation of the user's request, rather than merely containing isolated correct actions, a plausible start, or a correct-looking endpoint.
- The task-critical stages or milestones appear with sufficient visible specificity and in the required or logically necessary order, with no important omission, skipped dependency, premature outcome, unrelated segment, contradiction or regression. Correct start and end states do not compensate for a missing critical process stage.

SCORE 1 [CORE ALIGNMENT REALIZED WITH SECONDARY DEFECTS]

- The user's intended core process is visibly recognizable and every indispensable task-critical milestone is substantially realized, but the overall process still has secondary incompleteness or imprecision.
- Any missing, ambiguous, premature, out-of-order, interrupted, regressed, unrelated, or visually unsupported content is limited to a non-core stage or secondary detail whose failure does not prevent realization of the user's central intent.
- If even one indispensable task-critical milestone is absent, only approached or initiated, visually ambiguous, skipped, contradicted, carried out in a fundamentally wrong way, or claimed without direct visual support, assign
  SCORE 0 rather than SCORE 1.
- When all indispensable milestones are realized, prominent unrequested instructional text, captions, step labels, arrows, or guide graphics may still reduce the sequence to SCORE 1.

SCORE 0 [KEY MILESTONE NOT REALIZED]

- At least one indispensable task-critical milestone is absent, only approached or initiated, visually ambiguous, skipped, premature, out of order in a way that breaks the process, interrupted, regressed, contradicted, carried out in a fundamentally wrong way, or claimed without direct visual support.
- A milestone is indispensable when it is explicitly emphasized by the text instruction or HINTS, is necessary to realize the user's central intent, or is a required dependency for a later key state or result.
- Other correct milestones, a matching subject or setting, isolated relevant actions, a plausible start, or a correct-looking endpoint do not compensate for one uncompleted indispensable milestone.
- The sequence also receives SCORE 0 when it remains essentially static despite required change, or when its changes do not form a recognizable realization of the user's intent.

## Supplementary Rules

Base your judgment solely on the image itself; in most cases, any text or guiding cues added on top of the image should not affect your evaluation. When the image is only partially visible, evaluate it according to the information that is visible.

## Evaluation Steps

Use the text instruction, HINTS, and reference images to identify the task-critical milestones, mark as indispensable every milestone explicitly emphasized by the text instruction or HINTS, necessary to realize the user's central intent, or required as a dependency for a later key state or result, and identify any required or logically necessary ordering. Inspect the complete ordered sequence and assign the scores according to the Scoring Hierarchy. Assign an initial score using the scoring criteria. Then challenge that score by actively looking for previously missed problems, counterevidence, or alternative interpretations in the complete ordered sequence. Check whether these objections are supported by visible evidence and warrant changing the score; do not force a score change. Finally, give the revised final score (0, 1, or 2), retaining the initial score if it remains supported.

## Evaluation Data

- TEXT INSTRUCTION
- HINTS
- REFERENCE AND INITIAL IMAGES
- ALL ORDERED GENERATED IMAGES

## Output Contract

Return only a JSON object. Use an integer score (0, 1, or 2) and briefly explain the judgment with specific visual evidence.

{
  "score": 0,
  "evidence": "Brief visual evidence supporting the score."
}
\end{lstlisting}

\prompttitle{TC Sub-Metric: Outcome Completion (OC)}
\phantomsection
\label{app:oc-prompt}
\begin{lstlisting}[style=prompt]
# Outcome Completion Evaluation

Your task is to act as an OUTCOME COMPLETION AUDITOR in Strict Evaluation Mode. Given the text instruction, HINTS, reference images, and the final generated image, judge whether the final result fulfills the requested outcome. The original text instruction is for generating a sequence of frames, but no intermediate generated images are provided here, so you should only evaluate the final image and check if it meets the outcome of the instruction.

## Scoring Hierarchy

SCORE 1 [REQUESTED OUTCOME FULLY REALIZED]

- The final image visibly realizes the user's requested endpoint, and every indispensable outcome requirement is satisfied rather than merely approached, initiated, implied, or described.
- The required final entities, quantities, attributes, configuration, spatial relations, placement, and reference-specific features are correct wherever they are specified by the text instruction or HINTS or necessary to the user's central goal.
- The result is complete with sufficient visible specificity to verify it. No required final-state element is missing, contradictory, incorrectly substituted, or hidden in a way that prevents verification.

SCORE 0 [REQUESTED OUTCOME NOT FULLY REALIZED]

- At least one required endpoint condition or specified final-state detail is missing, incorrect, contradicted, incomplete, or visually unverifiable. Secondary defects in required details also receive SCORE 0; there is no partial credit.
- An outcome requirement is indispensable when it is explicitly emphasized as a required final condition by the text instruction or HINTS, or when its absence prevents realization of the user's central requested result. An intermediate milestone is not automatically an endpoint requirement.
- The outcome is only asserted by a caption, label, arrow, model-authored plan, or plausible narrative rather than shown. Do not infer successful placement, transfer, removal, or completion solely because an object is missing or obscured. When absence itself is the requested endpoint, credit it only if the relevant final region is visible and supports that requirement.

## Evaluation Steps

Read the text instruction, HINTS, and reference images to extract the requested endpoint. Separate final-state requirements from instructions about how to reach that state; do not require the final image to repeat intermediate actions. Inspect the final generated image and match each endpoint requirement to direct visible evidence. Assign an initial score using the scoring criteria. Then challenge that score by actively looking for previously missed problems, counterevidence, or alternative interpretations in the final generated image only. Check whether these objections are supported by visible evidence and warrant changing the score; do not force a score change. Finally, give the revised final score (0 or 1), retaining the initial score if it remains supported.

## Evaluation Data

- TEXT INSTRUCTION
- HINTS
- REFERENCE AND INITIAL IMAGES
- FINAL GENERATED IMAGE ONLY (ONE IMAGE; NO INTERMEDIATE GENERATED IMAGES)

## Output Contract

Return only a JSON object. Use an integer score (0 or 1) and briefly explain the judgment with specific visual evidence.

{
  "score": 0,
  "evidence": "Brief visual evidence supporting the score."
}
\end{lstlisting}

\prompttitle{Causal Plausibility (CP).}
\phantomsection
\label{app:cp-prompt}
\begin{lstlisting}[style=prompt]
# Causal Plausibility Evaluation

Your task is to act as a SEQUENCE-LEVEL CAUSAL AND WORLD-LAW AUDITOR in Strict Evaluation Mode. Inspect the complete ordered image sequence produced by the generation model and judge whether the overall process follows the object transformations, physical laws, game rules, everyday phenomena, logical dependencies, and task-world rules emphasized by the text instruction and HINTS. Apply realistic everyday physics by default, but follow explicit fictional, game, diagrammatic, symbolic, or other task-specific world rules when the task establishes them.

## Scoring Hierarchy

SCORE 3 [FULLY PLAUSIBLE]

- All concerns in the text instruction or HINTS are satisfied with direct visual evidence.
- The complete process is fully plausible, and every critical adjacent transition visibly follows from the state and action before it.
- The core mechanisms, constraints, object topology, and long-range dependencies all hold consistently, with no identifiable deviation from the governing rules and no skipped mechanism disguised by a plausible final state.

SCORE 2 [KEY MECHANISM PLAUSIBLE WITH OTHER MINOR FLAWS]

- Some new local rule deviation, weakened mechanism, incomplete long-range dependency, or minor unsupported transition is still found.
- The concerns in the text instruction or HINTS are partially satisfied, but the intended causal mechanism remains recognizable: the critical states and transitions broadly follow the expected governing rules.
- The overall process remains fluent and broadly instruction-aligned; no critical mechanism is skipped, contradicted, or replaced by an abrupt structural transformation.

SCORE 1 [CORE MECHANISM RECOGNIZABLE BUT VISIBLY FLAWED]

- The intended core causal mechanism remains visibly recognizable, and at least one critical cause-to-effect transition is supported, but a core concern in the text instruction or HINTS or a key verifiable condition in the HINTS is clearly violated at a critical state or transition.
- The broad phenomenon follows the applicable physical law, game rule, everyday phenomenon, or task-world rule, but a critical action, prerequisite, constraint, or transition is missing, unsupported, or mechanically incompatible.
- Alternatively, the process itself may be broadly plausible but lacks a strong connection to the reference images, text instruction, or HINTS.
- Mere thematic resemblance or a plausible final state is insufficient for
  SCORE 1 when the central causal mechanism is not visibly realized.

SCORE 0 [CORE MECHANISM ABSENT, CONTRADICTORY, OR IMPOSSIBLE]

- The central causal mechanism required to produce the main change is absent, reversed, contradicted, or replaced by an unsupported transformation, so the intended process is not causally recognizable as realized.
- The main result depends on a task-critical object, action, constraint, or state appearing, disappearing, teleporting, duplicating, or changing incompatibly without a visible or ordinarily inferable cause.
- The overall process is self-contradictory, or a key mechanism makes the main change impossible under the applicable world rules.
- A matching subject, broad phenomenon, or plausible-looking endpoint does not rescue a sequence whose core causal chain is missing or fundamentally invalid.

## Supplementary Rules

- If the output is static or changes too little, judge it according to the text instruction; when the task requires visible change, Causal Plausibility cannot exceed 1.
- If any task-critical transition cannot be causally reconciled from the visible states under the applicable world rules, Causal Plausibility cannot exceed 1. Ordinary unshown intermediate motion may be inferred only when the visible endpoints are mutually compatible and require no invented intervention, constraint change, or hidden state.
- Grade an abrupt or omitted transition by its causal severity rather than abruptness alone. A harmless compressed intermediate action needs no penalty when the visible endpoints remain mutually reachable; a local weakened or imprecise mechanism supports SCORE 2; an unsupported task-critical transition is capped at SCORE 1; and a self-contradictory or world-impossible core mechanism supports SCORE 0.
- If a task-relevant object or tool visible in one generated image is absent from a later image without a visible or ordinarily inferable removal, placement, consumption, transformation, occlusion, or exit from the frame, object conservation and causal continuity are violated and Causal Plausibility cannot exceed 1—even if the object is no longer needed after the action. If the unexplained disappearance removes the central mechanism, makes the main result causally unsupported, or is part of an impossible or self-contradictory core process, assign SCORE 0.

## Evaluation Steps

- Follow the visible changes from the first generated image to the last. First identify the applicable world assumptions and the task-critical states, constraints, and dependencies from the text instruction, HINTS, and reference images. Use realistic everyday physics by default, but follow explicit fictional, game, diagrammatic, symbolic, or other task-specific rules when provided.
- For every adjacent pair, identify the visible prior state, the visible action or cause, and the resulting state. Track every task-relevant object and tool across the pair, including where it remains, how it moves, and whether any appearance, disappearance, duplication, consumption, transformation, occlusion, or exit from the frame has visible or ordinarily inferable causal support. Inspect the relevant prerequisites and constraints, including support, contact, containment, object topology and component connectivity, material behavior, conservation of object count, temporal ordering, tool availability, geometry, and task-specific logical rules. Judge whether each cause can plausibly produce the state that follows.
- Check non-adjacent and long-range dependencies, including whether an early action enables or prevents a later state. Grade deviations by causal severity: distinguish harmless temporal compression, a local weakened or imprecise mechanism, an unsupported task-critical transition, and an impossible or self-contradictory core process. Keep Causal Plausibility separate from visual continuity: count an identity, shape, or appearance change here only when it breaks the causal mechanism or makes the resulting state unreachable under the applicable world rules.
- Assign an initial score using the scoring criteria. Then challenge that score by actively looking for previously missed problems, counterevidence, or alternative interpretations in the complete ordered sequence. Check whether these objections are supported by visible evidence and warrant changing the score; do not force a score change. Finally, give the revised final score (0, 1, 2, or 3), retaining the initial score if it remains supported.

## Evaluation Data

- TEXT INSTRUCTION
- HINTS
- REFERENCE AND INITIAL IMAGES
- ALL ORDERED GENERATED IMAGES

## Output Contract

Return only a JSON object. Use an integer score (0, 1, 2, or 3) and briefly explain the judgment with specific visual evidence.

{
  "score": 0,
  "evidence": "Brief visual evidence supporting the score."
}
\end{lstlisting}

\prompttitle{Visual Consistency (VC).}
\phantomsection
\label{app:vc-prompt}
\begin{lstlisting}[style=prompt]
# Visual Consistency Evaluation

Your task is to act as a MULTI-FRAME CONTINUITY AUDITOR in Strict Evaluation Mode. Compare the complete ordered image sequence and evaluate whether visual state that should persist remains consistent across frames. Only the specific state change explicitly requested by the user or visibly required by the causal process is exempt from Visual Consistency penalties; unrelated changes to identity, geometry, material, topology, mechanism structure, camera, or background remain Visual Consistency errors.

## Scoring Hierarchy

SCORE 3 [FULLY CONSISTENT]

- Task-relevant invariants such as identity, count, shape, scale, color, material, topology, mechanism structure, relative position, camera, and background remain stable.
- Every change is directly justified by the visible process, with no identifiable appearance, disappearance, duplication, teleportation, structural replacement, or cumulative drift.
- Unless explicitly requested, no new instructional caption, step label, explanatory text, arrow, or guide graphic appears in the generated images; scene-native text already present in the reference is preserved normally.

SCORE 2 [MINOR ISSUES BUT KEY OBJECTS STABLE]

- Key objects emphasized by the HINTS or text instruction show only slight drift that can recover in later frames.
- Objects not emphasized by the HINTS or text instruction may have a small but real identity, geometry, scale, position, or background continuity deviation.

SCORE 1 [CONTINUITY ISSUE BUT KEY OBJECTS RECOGNIZABLE]

- The target remains broadly recognizable, but continuity is poor; a key object has at least one clear color, identity, geometry, scale, topology, mechanism-structure, position, or background continuity deviation.
- Objects not emphasized by the HINTS or text instruction have strong deviations that affect the whole sequence.

SCORE 0 [TOTAL FAILURE]

- The target object is not preserved at all, or there is unexplained loss, addition, or duplication of important objects, a major scale jump, teleportation, background reset, severe deformation, topology or mechanism replacement, bizarre texture, or inexplicable distortion.
- An important object disappears without a visible removal or placement action. Do not infer that it was safely moved merely because it is absent later.
- If an object or tool explicitly named by the HINTS, or required to remain consistent in instruction disappears, appears, or duplicates without a visible removal, placement, entry, exit, consumption, transformation, or occlusion, assign SCORE 0.

## Evaluation Steps

- Compare adjacent and non-adjacent images across the complete ordered sequence. Track every task-relevant object and tool from frame to frame, including its count and visible location, and form a score by checking for unjustified changes in identity, count, shape, scale, color, material, topology, mechanism structure, relative position, camera, or background, including appearance, disappearance, duplication, teleportation, structural replacement, cumulative drift, distortion. Exempt only the specific visual changes directly required by the text instruction, or visibly established causal process.
- Assign an initial score using the scoring criteria. Then challenge that score by actively looking for previously missed problems, counterevidence, or alternative interpretations in the complete ordered sequence. Check whether these objections are supported by visible evidence and warrant changing the score; do not force a score change. Finally, give the revised final score (0, 1, 2, or 3), retaining the initial score if it remains supported.

## Evaluation Data

- TEXT INSTRUCTION
- HINTS
- REFERENCE AND INITIAL IMAGES
- ALL ORDERED GENERATED IMAGES

## Output Contract

Return only a JSON object. Use an integer score (0, 1, 2, or 3) and briefly explain the judgment with specific visual evidence.

{
  "score": 0,
  "evidence": "Brief visual evidence supporting the score."
}
\end{lstlisting}

\prompttitle{Designated-Step Accuracy.}
\phantomsection
\label{app:designated-step-accuracy}
This evaluator is used for the main-text analysis
\emph{Interleaved generation faces the challenge of accumulated context} in
Section~\ref{sec:generation-schema-analysis}. It measures how generation
accuracy changes across successive steps as interleaved context accumulates.
\par\smallskip
\begin{lstlisting}[style=prompt]
# Step Accuracy Evaluation

Your task is to act as a DESIGNATED-STEP ACCURACY AUDITOR in Strict Evaluation Mode. Use the text instruction, HINTS, reference images, and the model's text plan paired with the designated generated image to evaluate that image only. Extract the concrete milestone or task-relevant change required for this step, then judge whether the image visibly realizes that requirement, makes positive task-relevant progress, and matches its paired plan.

## Scoring Hierarchy

SCORE 1 [STEP REQUIREMENT REALIZED]

- The image visibly realizes the requirement for this exact step in the ordered sequence. Any explicitly required action, state, entity, quantity, attribute, spatial relation, or reference relation for this step is satisfied.
- The visible action or state agrees with the model's text plan paired with this image and produces positive progress toward the user's task. The model's plan does not override the user's instruction, HINTS, or reference requirements; following a wrong or irrelevant plan is not sufficient.
- The designated step satisfies the constraints applicable to its own action or state and is physically plausible. Required contact, support, containment, geometry, object count, and material behavior are visibly compatible with the depicted action.
- No task-relevant object appears, disappears, duplicates, teleports, or abruptly deforms without a task-valid explanation supported by the supplied visual evidence. Legitimate task-required transformations and supported occlusion are not errors.
- Requirements in HINTS that are declared explicitly relevant to this step are satisfied.

SCORE 0 [STEP REQUIREMENT NOT REALIZED]

- The visible change is irrelevant to constitute the required milestone. A repeated or essentially unchanged image fails unless maintaining that state is explicitly required for this step.
- The designated image omits or violates a required action, state, entity, quantity, attribute, spatial relation, reference relation, or local constraint for this step; uses the wrong reference; or depicts a different action or state from the one required for this step.
- The image contradicts its paired model-authored plan, or the planned and depicted action fails to produce task-relevant positive progress. Do not treat the plan's claim of success as evidence that the image succeeded.
- The designated action or state has a visible physical or task-world violation, such as impossible contact or support, incompatible geometry, or an unexplained appearance, disappearance, duplication, teleportation, or abrupt deformation of a task-relevant object.
- Requirements in HINTS that are declared explicitly relevant to this step are not satisfied or only partially satisfied.

## Evaluation Steps

### Identify the Designated Step

- Read the designated step index, original text instruction, HINTS, reference images, and model-authored text plan paired with the designated image. Evaluate only that step; do not produce judgments for other steps.
- Extract the concrete action or state promised for this step, its task-relevant visual requirements, and relevant HINTS. The user's task takes priority over an incorrect model plan.

### Evaluate the Designated Generated Image

- Inspect the designated image's action, entities, quantities, attributes, and relations. Check that it substantially realizes its own step requirement and makes meaningful positive progress.
- Compare the image with its paired textual plan, then check local physical plausibility and task-relevant object integrity. Any deduction must cite visible evidence attributable to the designated step.
- Other supplied images are context only: do not penalize the designated step because another step fails. Do not invent an unseen comparison state to claim disappearance or deformation.

### Recheck and Return One Judgment

- Reinspect the designated image and challenge the initial judgment against both score definitions. Correct any unsupported inference or cross-step blame. Output the accuracy in SCORE 0 or 1.

## Evaluation Data

- TEXT INSTRUCTION
- HINTS
- REFERENCE AND INITIAL IMAGES
- DESIGNATED STEP INDEX
- MODEL-AUTHORED TEXT PLAN PAIRED WITH THE DESIGNATED IMAGE
- DESIGNATED GENERATED IMAGE

## Output Contract

Return only a JSON object. Use an integer score (0 or 1) and briefly explain the judgment with specific visual evidence. Set step_index to the designated step; do not score other steps.

{
  "step_index": 1,
  "score": 0,
  "evidence": "Brief visual evidence for the designated step."
}
\end{lstlisting}

\subsection{Agentic Generation Harnesses}
\label{app:agentic-prompts}

\prompttitle{Inspection Loop without Coding Tools.}
\begin{lstlisting}[style=prompt]
Tool Configuration: code_mode: off; shell: off; web_search: disabled
\end{lstlisting}
\begin{lstlisting}[style=prompt]
Main Prompt:
{{ORIGINAL_TASK}}

## Required Workflow

1. Plan the intended image sequence using the task and supplied references.
2. Generate the images using the image-generation tool or API assigned to the current baseline.
3. Inspect each image and the full sequence for task compliance, geometry, orientation, and continuity. Edit or regenerate when useful, then inspect again.
4. Select the best existing sequence within the shared call and time budgets. Do not revise images that are already correct.

## Constraints

- Use the image-generation and editing tool or API assigned to the current baseline; viewing images is unrestricted. Do not switch to another image model or provider.
- Do not use code or modeling tools to construct, render, or use intermediate visual artifacts.
- All image-generation and editing attempts, including failures, share one per-Case budget of 2 × {{STEP_COUNT}} calls ({{MAX_IMAGE_CALLS_2N}} submissions). Initial generations, regenerations, and edits all count. Never reset or bypass this budget.
- Non-image-generation tool use does not consume image-generation calls, but must remain within the Case time limit. Any image-generation or editing API call counts regardless of its purpose.
- At the cap, make no further image-generation or editing calls. Select from existing outputs and report any remaining shortfall.
- You have 150 minutes total, with reminders every 15 minutes.
- Use the supplied task and references and your own work from this Case. Do not read other Cases, sessions, baseline results, scores, Judge hints, or evaluation files. Do not modify source data, reference files, other runs' outputs, the budget gate, ledger, or configuration, or start other sessions.

## Output Contract

- Return exactly {{STEP_COUNT}} accepted final images in step order, one
  absolute Markdown image link per step:
  ![Step N](/absolute/path/to/image.png).
- Include only the selected final images, not discarded drafts or superseded revisions. Verify that each path points to an existing image.
- If the required sequence cannot be completed within the budget, return the available selected images and briefly state the shortfall; do not invent files or exceed the budget.
\end{lstlisting}

\prompttitle{Inspection Loop Encouraging Coding Tools.}
\begin{lstlisting}[style=prompt]
Tool Configuration: code_mode: on; shell: on; web_search: disabled
\end{lstlisting}
\begin{lstlisting}[style=prompt]
Main Prompt:
{{ORIGINAL_TASK}}

Each step contains one textual instruction and one image. Don't output all steps in one image, instead output {{STEP_COUNT}} images interleaved with text description. Do not render the textual instructions, step labels, captions, explanatory text, arrows, or guide graphics inside the images unless visible text is explicitly required by the user's task.

## Required Workflow

1. Plan each intended output. Use the task and references to describe the content, composition, action or state, viewpoint, and continuity requirements of every image. Check that the descriptions form the requested progression and distinguish intended changes from properties that must remain stable. Gather any additional context from the supplied task and references when useful.
2. Build a task-specific visual reasoning artifact before the first image-generation call. Use code or a modeling tool to represent the relevant geometry, arrangement, state progression, motion, or causal relationships. Choose a suitable method, such as SVG, Canvas, Three.js, Python, or another renderer; no language is required. Render a meaningful visual result, not just code, logs, or prose.
3. Open and visually inspect the rendering. Check assumptions, spatial relationships, state transitions, and task constraints against what is actually visible. Correct empty, misleading, or inaccurate artifacts and inspect them again. Use the findings to refine the image descriptions. A successful script exit alone is not visual inspection.
4. Prepare and generate the sequence. Consolidate the descriptions, references, and artifact findings into clear generation instructions. Explicitly decide whether the inspected artifact should be supplied as an image reference; attaching it is optional, but constructing and inspecting it are required. Use the image-generation and editing tool or API assigned to the current baseline.
5. Inspect and revise. Examine every generated image and the complete ordered sequence for task compliance, direction, geometry, causal plausibility, visual continuity, and the final outcome. When useful, revise the plan, supporting context, code artifact, or generated images. Render and inspect changed artifacts before using them, and recheck the sequence after each material revision. Do not revise outputs that are already correct.
6. Select the final sequence. Stop within the shared call and time budgets, choose the best existing outputs, and verify their order and quantity. State unresolved limitations rather than bypassing the limits.

## Constraints

- Use the image-generation and editing tool or API assigned to the current baseline; viewing images is unrestricted. Do not switch to another image model or provider.
- You may use code or modeling tools to construct, render, inspect, and revise intermediate visual artifacts.
- All image-generation and editing attempts, including failures, share one per-Case budget of 2 × {{STEP_COUNT}} calls ({{MAX_IMAGE_CALLS_2N}} submissions). Initial generations, regenerations, and edits all count. Never reset or bypass this budget.
- Non-image-generation tool use does not consume image-generation calls, but must remain within the Case time limit. Any image-generation or editing API call counts regardless of its purpose.
- At the cap, make no further image-generation or editing calls. Select from existing outputs and report any remaining shortfall.
- You have 150 minutes total, with reminders every 15 minutes.
- Use the supplied task and references and your own work from this Case. Do not read other Cases, sessions, baseline results, scores, Judge hints, or evaluation files. Do not modify source data, reference files, other runs' outputs, the budget gate, ledger, or configuration, or start other sessions.

## Artifact Records

Save the source and rendered visual files in the Case workspace. Record source_path, rendered_path, render_command, and inspection_finding in visual_artifacts.json, along with whether the artifact is used as generation context. For interactive or animated artifacts, save representative frames or a video. Preserve useful versions and revision history separately from final outputs.

Bare code, JSON, console output, prose screenshots, unchanged references, copies of generated images, and artifacts created only after image generation do not satisfy the pre-generation requirement. If an artifact does not help resolve uncertainty or validate the plan, improve it before proceeding. Do not modify source data, reference files, or other runs' outputs.

## Output Contract

- Return exactly {{STEP_COUNT}} accepted final images in step order, interleaved with text, using absolute Markdown image paths.
- Include only selected final images, not code artifacts, discarded drafts, or superseded revisions. Verify that the image files exist and that the output order and count are correct.
- If the required sequence cannot be completed within the budget, return the available selected images and briefly state the shortfall; do not invent files or exceed the budget.
\end{lstlisting}

\fi
\end{document}